%% file: main.tex
\documentclass[10pt,journal,compsoc]{IEEEtran}
\ifCLASSOPTIONcompsoc
  \usepackage[nocompress]{cite}
\else
  \usepackage{cite}
\fi

\ifCLASSINFOpdf
\else
\fi

\usepackage{amsmath,amssymb,amsfonts}
\usepackage{graphicx}
\usepackage{textcomp}
\usepackage{xcolor}
\def\BibTeX{{\rm B\kern-.05em{\sc i\kern-.025em b}\kern-.08em
    T\kern-.1667em\lower.7ex\hbox{E}\kern-.125emX}}

\usepackage{tikz}

\usepackage{filecontents}

\usepackage{multirow}
\usepackage{textcomp}
\usepackage{balance}
\usepackage{array}

\usepackage{subfig}
\usepackage{algorithm}
\usepackage[noend]{algpseudocode}

\usepackage{dashrule}
\usepackage{url}

\usepackage{multirow}
\usepackage{hhline}
\usepackage{booktabs}
\usepackage{threeparttable}
\usepackage[shortlabels]{enumitem}
\usepackage{bm}
\usepackage{color}

\usepackage{ragged2e}
\usepackage{tcolorbox}

\begin{document}
%
\title{RNN-Test: Towards Adversarial Testing for Recurrent Neural Network Systems}
%
%
%
%

\author{Jianmin~Guo,
	    Yue~Zhao,
	    Quan~Zhang,
        and Yu~Jiang
\IEEEcompsocitemizethanks{
\IEEEcompsocthanksitem{J. Guo, Q.~Zhang and Y. Jiang are with School of Software, Tsinghua University, Beijing National Research Center for Information Science and Technology, and Key Laboratory for Information System Security, Ministry of Education, Beijing, 100084, China. 
}
\IEEEcompsocthanksitem{Y. Zhao is with Huawei Technologies Co., Ltd.}
\IEEEcompsocthanksitem{J. Guo is corresponding author. Email: guojm17@mails.tsinghua.edu.cn.}
}
\thanks{Manuscript submitted XX XX, 2021.}}

\markboth{Journal of \LaTeX\ Class Files,~Vol.~xx, No.~xx, August~20xx}%
{Guo \MakeLowercase{\textit{et al.}}: RNN-Test: Towards Adversarial Testing for Recurrent Neural Network Systems}
%



\IEEEtitleabstractindextext{%
\begin{abstract}
\input{sec-abstract}
\end{abstract}

}

\maketitle

\IEEEdisplaynontitleabstractindextext

%
\IEEEpeerreviewmaketitle

\input{sec-introduction}
\input{sec-background}
\input{sec-coverage}
\input{sec-approach}

\input{sec-experiment}
\input{sec-discussion}
\input{sec-relatedwork}
\input{sec-conclusion}

\ifCLASSOPTIONcaptionsoff
  \newpage
\fi




\bibliographystyle{IEEEtran}
\bibliography{sample-references}

\begin{IEEEbiography}[{\includegraphics[width=1in,height=1in,clip,keepaspectratio]{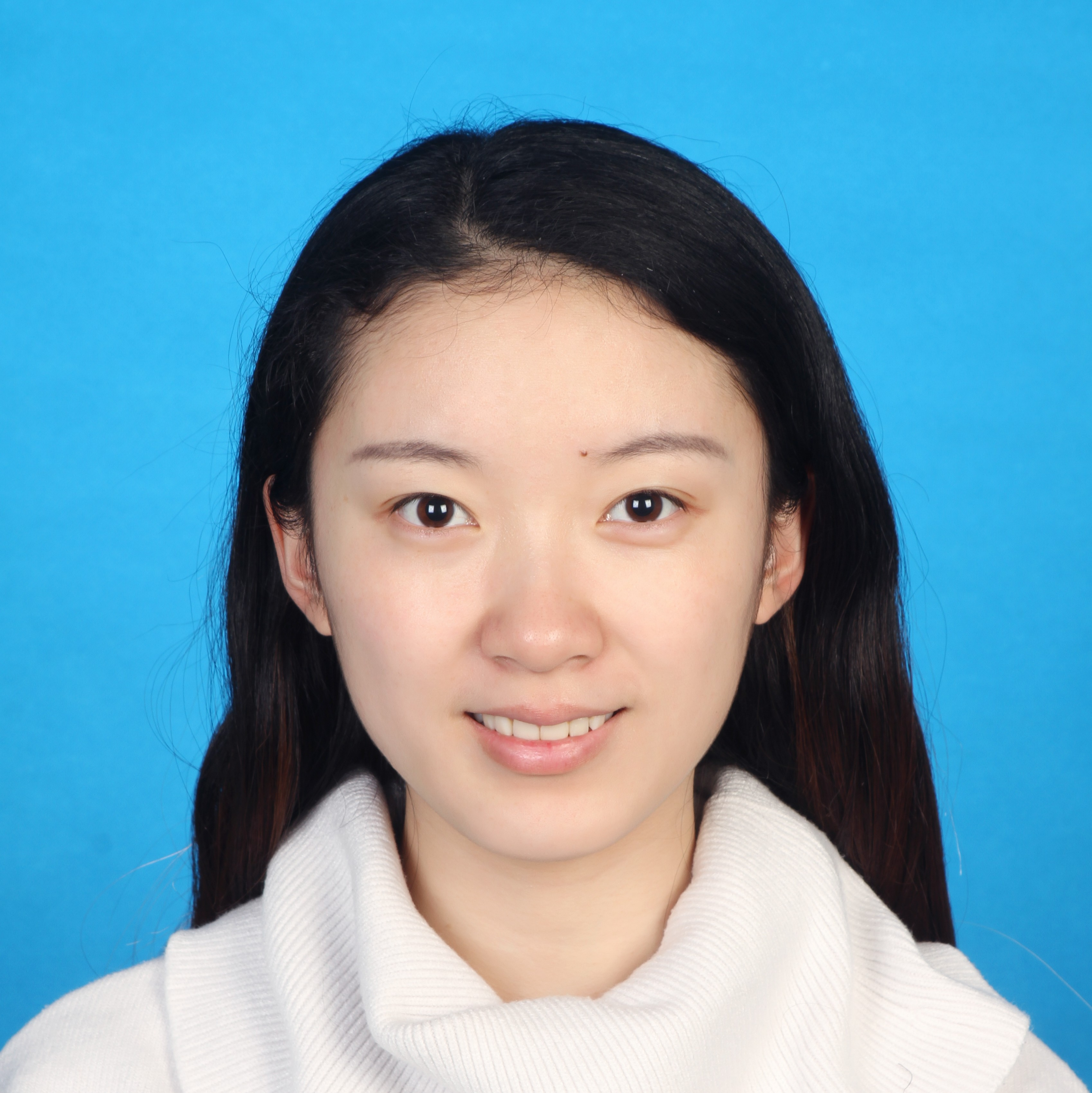}}]{Jianmin Guo}
is a Ph.D. candidate at School of Software Engineering, Tsinghua University, Beijing, China. She received the BS degree in software engineering from Beijing University of Posts and Telecommunications, Beijing, China, in 2017. Her research interests are software testing, mainly focusing on deep learning testing and adversarial testing of recurrent neural networks.
\end{IEEEbiography}

\vspace{-1cm}
\begin{IEEEbiography}[{\includegraphics[width=1in,height=1in,clip,keepaspectratio]{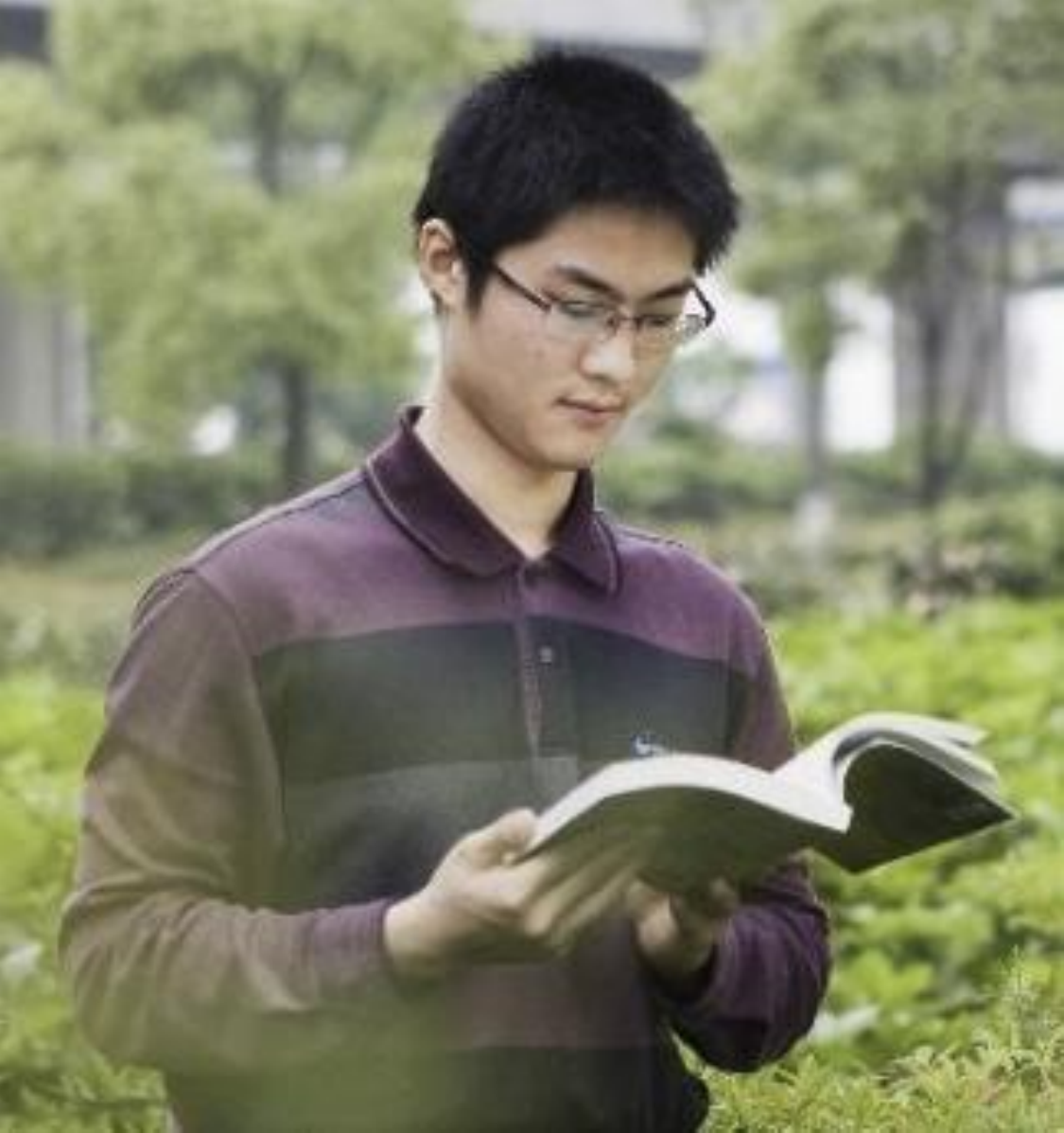}}]{Yue Zhao}
is a software engineer in Huawei Technologies Co., Ltd. He received MS degree at School of Software Engineering, Tsinghua University, Beijing, China, in 2020. He received the BS degree in software engineering from Beijing University of Posts and Telecommunications, Beijing, China, in 2017. His research interests are deep learning testing and backdoor detection of deep learning systems.
\end{IEEEbiography}

\vspace{-1cm}
\begin{IEEEbiography}[{\includegraphics[width=1in,height=1in,clip,keepaspectratio]{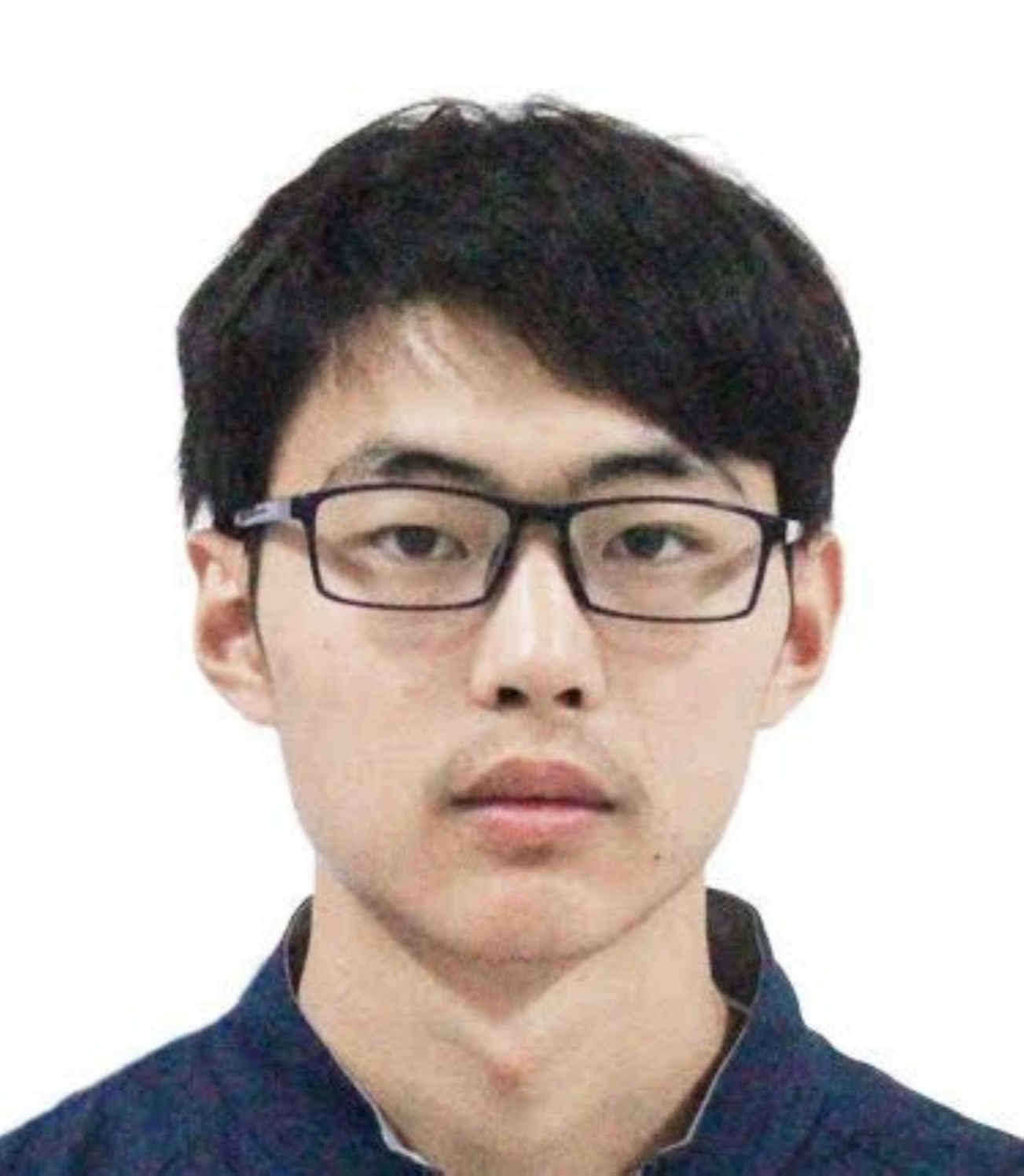}}]{Quan Zhang}
is a Ph.D. student at School of Software Engineering, Tsinghua University, Beijing, China. He received the BS degree in computer science from Beijing University of Posts and Telecommunications, Beijing, China, in 2020. His research interests are backdoor detection of deep learning systems.
\end{IEEEbiography}

\vspace{-1cm}
\begin{IEEEbiography}[{\includegraphics[width=1in,height=1in,clip,keepaspectratio]{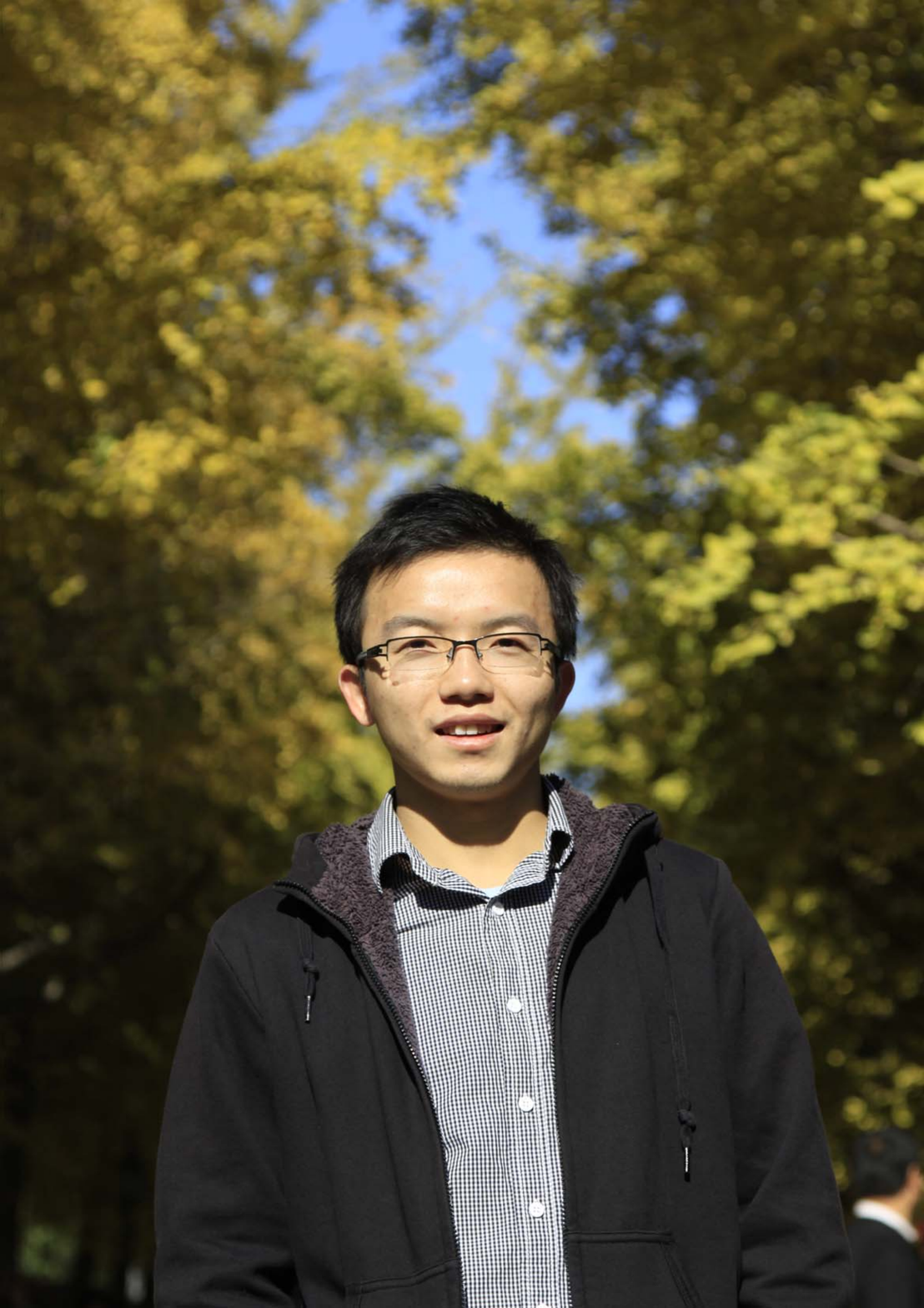}}]{Yu Jiang}
received the BS degree in software engineering from Beijing University of Posts and Telecommunications, Beijing, China, in 2010, and the PhD degree in computer science from Tsinghua University, Beijing, China, in 2015. He was a Postdoc researcher in the department of computer science of University of Illinois at Urbana-Champaign, IL, USA, in 2016. He is now an associate professor in Tsinghua University, Beijing, China. His current research interests include domain specific modeling, formal computation model, formal verification and their applications in embedded
systems. 
\end{IEEEbiography}

\end{document}

%% file: sec-abstract.tex
\justifying	
While massive efforts have been investigated in adversarial testing of convolutional neural networks (CNN), testing for recurrent neural networks~(RNN) is still limited and leaves threats for vast sequential application domains. 
%
In this paper, we propose an adversarial testing framework RNN-Test for RNN systems, focusing on the main sequential domains, not only classification tasks. 
First, we design a novel search methodology customized for RNN models by maximizing the inconsistency of RNN states to produce adversarial inputs. Next, we introduce two state-based coverage metrics according to the distinctive structure of RNNs to explore more inference logics. Finally, RNN-Test solves the joint optimization problem to maximize state inconsistency and state coverage, and crafts adversarial inputs for various tasks of different kinds of inputs.


For evaluations, we apply RNN-Test on three sequential models of common RNN structures. On the tested models, the RNN-Test approach is demonstrated to be competitive in generating adversarial inputs, outperforming FGSM-based and DLFuzz-based methods to reduce the model performance more sharply with 2.78\% to 32.5\% higher success~(or generation) rate. RNN-Test could also achieve 52.65\% to 66.45\% higher adversary rate on MNIST-LSTM model than relevant work testRNN. Compared with the neuron coverage, the proposed state coverage metrics as guidance excel with 4.17\% to 97.22\% higher success~(or generation) rate. 

%% file: sec-introduction.tex
\IEEEraisesectionheading{\section{Introduction}\label{sec:introduction}}
\IEEEPARstart{A}{s} the core part of the current artificial intelligence applications, deep learning has made great breakthroughs in computer vision~\cite{Krizhevsky2012ImageNet}, natural language processing~(NLP)~\cite{Collobert:2008:UAN:1390156.1390177}, and automatic speech recognition~(ASR)~\cite{speech-recognition}. With the increasing deployments of deep neural network~(DNN) systems in the safety- and security-critical domains, such as autonomous driving~\cite{bojarski2016end} and medical diagnose~\cite{shen2017deep}, ensuring the robustness of DNNs becomes an essential concern in both academic research and security communities. 

However, it is demonstrated that state-of-the-art DNN systems~\cite{szegedy2013intriguing} are easy to suffer attacks and produce completely wrong predictions, when fed with adversarial inputs which are nearly indistinguishable from original test inputs. This inspired numerous adversarial testing works devoted to generating adversarial inputs for DNNs, aiming to provide rich sources to train the DNNs to be more robust. The majority of these works~\cite{Goodfellow2015Explaining, papernot2016limitations, moosavi2016deepfool} try to fool popular image classifiers by applying minute perturbations to the inputs. They exhibit high efficiency but achieve low testing completeness~\cite{pei2017deepxplore}. Recently, researchers\cite{pei2017deepxplore, sun2018testing, ma2018deepgauge} try to apply traditional software testing techniques over DNNs, with various neuron-based coverage criteria proposed to measure the testing completeness of DNN logics. These works could reach high testing coverage and produce numbers of adversarial inputs.

In spite of the efficiency of these works, they are largely limited to CNNs and image classification tasks. Overall, there are two main types of DNN, the convolutional neural networks~(CNN)~\cite{lecun1995convolutional} and recurrent neural networks~(RNN)~\cite{rodriguez1999recurrent}. They are of distinct structures and preferred for different kinds of tasks. CNN exhibits excellent competence in dealing with image processing tasks~\cite{simonyan2014very, he2016deep}, with thousands of neurons good at extracting image features. RNN is known for the iterative structure over cells and specific components dealing with context semantics, hence expert in handling tasks with sequential data, like natural language processing~\cite{mikolov2011extensions} and speech recognition\cite{graves2013speech}. Owing to the huge gap, the testing techniques and coverage metrics for the two types of DNNs are hard to fit the other. 

So far, adversarial testing for RNN systems has received limited attention, especially those of sequential tasks~(outputs without explicit class labels). Existing works~\cite{sato2018interpretable, samanta2017towards, du2019deepstellar, huang2019coverage} still concentrate more on classification domains, thanks to advances in classification tasks of CNNs. They perform well over tasks such as sentiment analysis~\cite{sato2018interpretable, samanta2017towards,huang2019coverage}, image classification~\cite{du2019deepstellar, huang2019coverage}, and lipophilicity prediction~\cite{huang2019coverage}, etc. 
But the core sequential tasks of RNNs leave tested insufficiently, threatening their large-scale applications. 
Moreover, existing coverage criteria~\cite{pei2017deepxplore, sun2018testing, ma2018deepgauge} are mostly designed for CNNs and neurons, with a large gap to fit for RNNs. Lastly, recent works~\cite{du2019deepstellar, huang2019coverage} also proposed coverage guided testing methods with specific coverage criteria for RNN systems. 
They could generate lots of adversarial inputs for tested models~(mainly of classification tasks), by mutating inputs directly~(e.g. random noise) and employing coverage values as constraints to terminate testing. However, the specific inference logics of RNNs are not utilized in searching for adversarial inputs. 

Therefore, challenges for RNN testing are summarized as threefold. 
First, adversarial testing methods for RNNs with the main sequential domains are rather inadequate, leaving threats for majority application scenarios. 
For tasks with sequential outputs, there is no standard to recognize the inputs as adversarial inputs without obvious class labels.
Second, neuron-based coverage metrics fail to consider characteristics of RNN structures and could not be adopted directly. Third, existing testing methods are limited in making use of distinct logics of RNN models.

\vspace{0.1cm}
\noindent\textbf{Approach: }In this paper, we propose an adversarial testing framework RNN-Test for RNN systems, especially those with sequential outputs. RNN-Test concentrates on the kernel structure for sequential processing and rids of remain structures for particular applications. According to the unique features of RNNs, we put forward a specific search methodology, which maximizes the inconsistency of RNN states to obtain adversarial inputs. Meanwhile, we also design two state-based coverage metrics for different RNN models to explore more logics and guide to discover adversarial inputs in irregular space. They are then combined as a joint optimization problem, which is to maximize the state inconsistency and state coverage. Finally, it will be solved to acquire perturbations in a gradient-based way. 

When obtained the perturbations, adversarial inputs will be crafted by applying perturbations to original test inputs in different ways for various kinds of inputs. In the end, we adopt model performance metrics to identify the adversarial inputs and assess their qualities, which are generally available for RNN variants. 
In this way, RNN-Test provides a scalable and extensible solution for RNN testing. 

\vspace{0.1cm}
\noindent\textbf{Evaluation: }We evaluate the RNN-Test approach over three RNN models of sequential tasks, including two NLP models with discrete inputs~(PTB language model~\cite{ptb}, spell checker model~\cite{spellchecker}) and one ASR model with continuous inputs~(DeepSpeech ASR model~\cite{hannun2014deep}). The RNN-Test approach could efficiently acquire adversarial inputs of high quality, which reduce the model performance sharply while nearly imperceptible to original inputs. 
Compared with baselines which are two popular techniques~(FGSM~\cite{Goodfellow2015Explaining} and DLFuzz~\cite{guo2018dlfuzz}) adapted here for RNN testing, our approach achieves more performance reduction with higher success~(or generation) rate. Taking DeepSpeech ASR model as an example, RNN-Test could decline the model performance by 17.29\% higher WER, 3.61\% lower BLEU with 10\% higher success rate, in contrast with the FGSM-based method.
As for the most relevant work testRNN~\cite{huang2019coverage}, RNN-Test could achieve 52.65\% to 66.45\% higher adversary rate on MNIST-LSTM model~\cite{mnist-lstm}.

Furthermore, coverage guidance as pure optimization is first demonstrated with diverse searching capability for adversarial inputs compared with other methods. 
The proposed state coverage guidance achieved 4.17\% to 97.22\% higher success~(or generation) rate than neuron coverage guidance, and even best performance on the spell checker model. 
Finally, adversarial inputs obtained by RNN-Test are of high quality. Besides reducing the model performance sharply with minute perturbations and high time efficiency, they could also improve the model by retraining, such as 12.582\% improvement of test perplexity on PTB language model.

\vspace{0.1cm}
\noindent\textbf{Contribution: }Our work has the following contributions:
\begin{itemize}
	\item We design a novel search methodology based on the inner logics of RNNs, which maximizes the inconsistency of RNN states to produce adversarial inputs efficiently.
    \item We propose two state-based coverage metrics customized for RNNs, mainly as guidance for adversarial testing. We first demonstrate that coverage guidance has a diverse searching capability for adversarial inputs compared with other methods. Our state coverage guidance is also superior to neuron coverage guidance in RNN testing. 
    \item We design and implement the adversarial testing framework RNN-Test. To our best knowledge, it is the first step towards systematically testing the majority sequential domains for RNN systems, referred to as tasks of sequential outputs without class labels. It is effective and scalable for variants of RNNs, outperforming FGSM- and DLFuzz-based methods as well as testRNN~(on MNIST-LSTM model). 

\end{itemize}


%% file: sec-background.tex
\section{Background}\label{sec:background}
\subsection{Deep Neural Network} 
In the following, we will describe the two main kinds of DNN, convolutional neural networks~(CNN)~\cite{lecun1995convolutional} and recurrent neural networks~(RNN)~\cite{rodriguez1999recurrent}.

\begin{figure}[!htbp]
    \centering
        \subfloat[Typical CNN structure]{\includegraphics[width=0.67\linewidth]{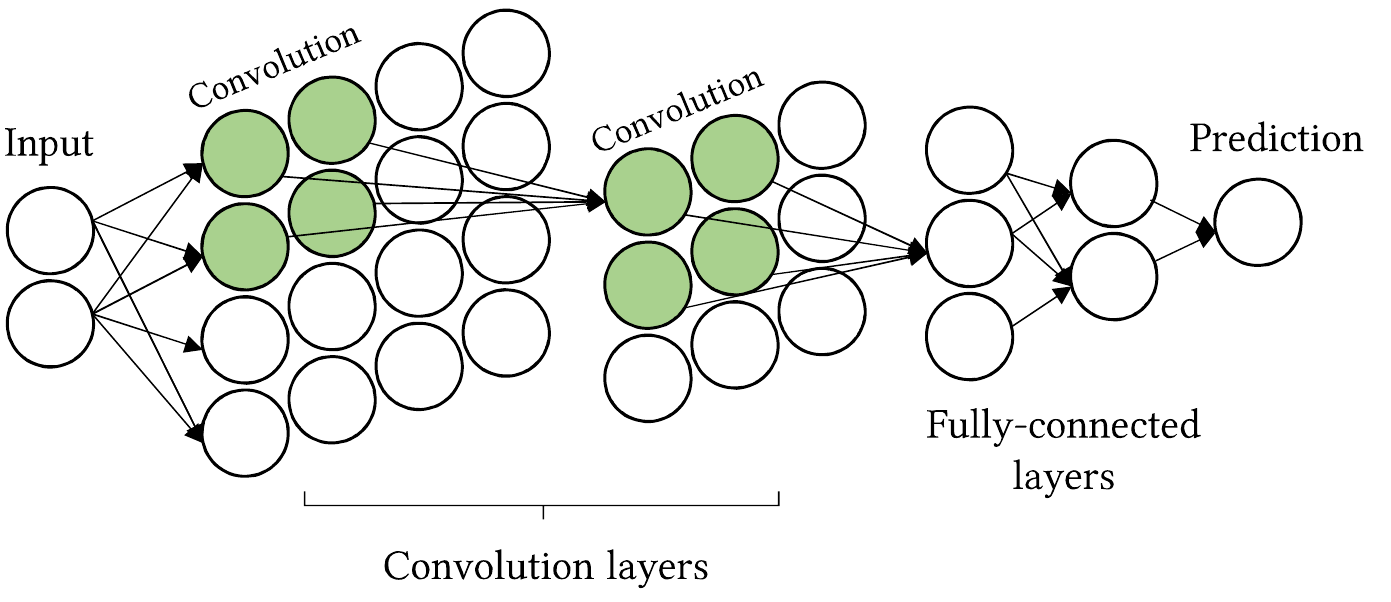}%
         \label{fig:traditional DNN}}
        \subfloat[Classical neuron]{\includegraphics[width=0.31\linewidth]{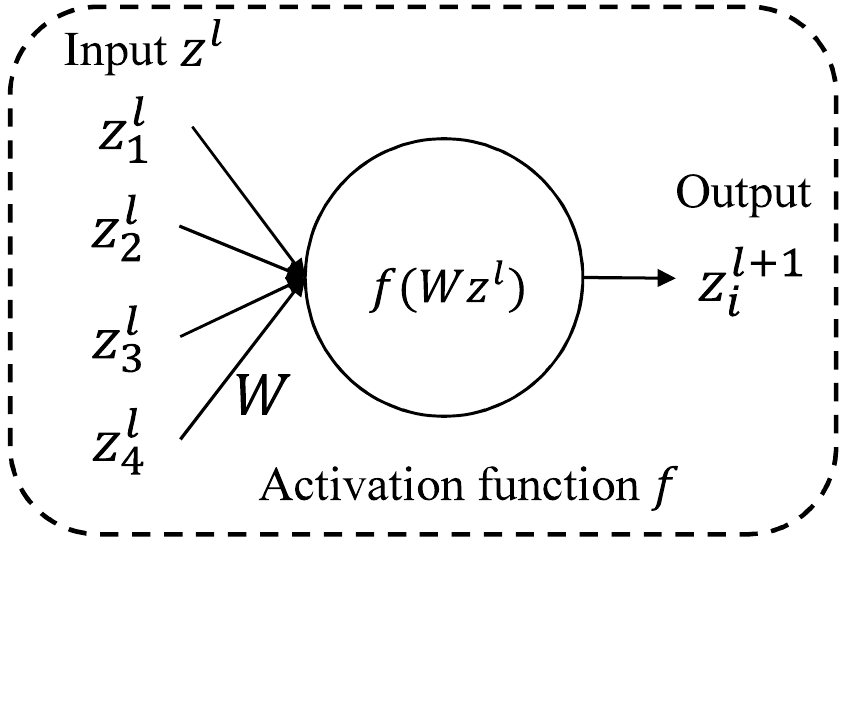}%
         \label{fig:Typical neuron}}
         \vspace{-0.1cm}
    \caption{CNN structure and neuron} 
    \label{fig:DNN structure}
\end{figure}

\textbf{Convolutional neural network and neuron.} 
Fig.~\ref{fig:traditional DNN} shows the simplified structure of a typical CNN. CNN keeps the fundamental feed-forward structure, where each neuron is connected with neurons of adjacent layers while no connections with those of the same layer. They are broadly used in image processing tasks~\cite{simonyan2014very, he2016deep}, with specific convolution layers good at extracting image features. Besides, CNN adopts classical DNN neurons, as shown in Fig. \ref{fig:Typical neuron}. The neuron output is a single value transformed by a non-linear activation function, which is usually RELU~(Rectified Linear Unit).

\begin{figure}[!htbp]
    \centering
    \includegraphics[width=0.3\textwidth]{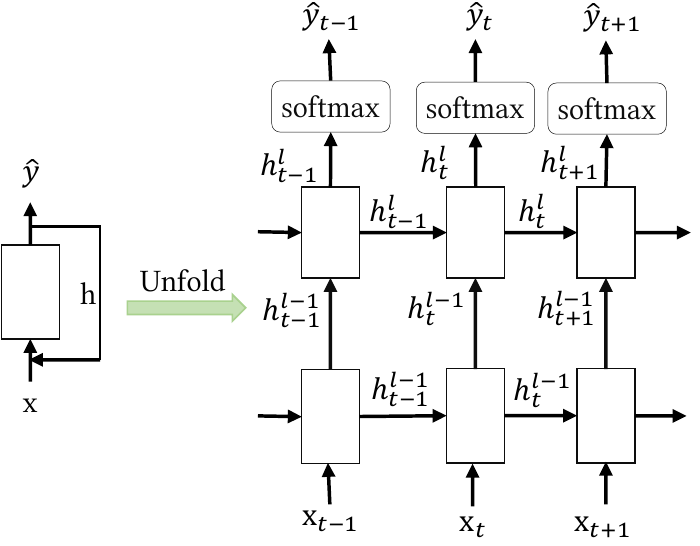}
    \vspace{-0.2cm}
    \caption{RNN structure}
    \label{fig:RNN structure}
\end{figure}

\textbf{Recurrent neural network and cell.} 
Fig.~\ref{fig:RNN structure} illustrates the basic structure of RNN, where an elementary RNN cell~(noted as the square) iteratively makes predictions $\hat{y}$ based on inputs $x$ and intermediate outputs $h$, which are referred to as hidden states. When it is unfolded, it can be seen that the input sequence $x$ is fed to the RNN cell as a series of time steps, where $x$ could be a sentence and $x_t$ is the $t$-th word. Moreover, each prediction $\hat{y_t}$ could be the predicted word right after $x_t$ based upon the received sequence.

In contrast to CNN neuron, the hidden state output $h_{t}^{l}$ of the cell at time step $t$ of layer $l$ is decided by current input $h_{t}^{l-1}$ from the previous layer as well as $h_{t-1}^{l}$ from the prior step in the same layer, and then passed forward to compute the softmax predictions. Due to this key design, RNN excels in making use of the interior contextual semantics of sequential inputs. 

\begin{figure}[!htbp]
\setlength{\belowcaptionskip}{-5pt}
    \centering    
    	\subfloat[General RNN cell]{\includegraphics[width=0.48\linewidth]{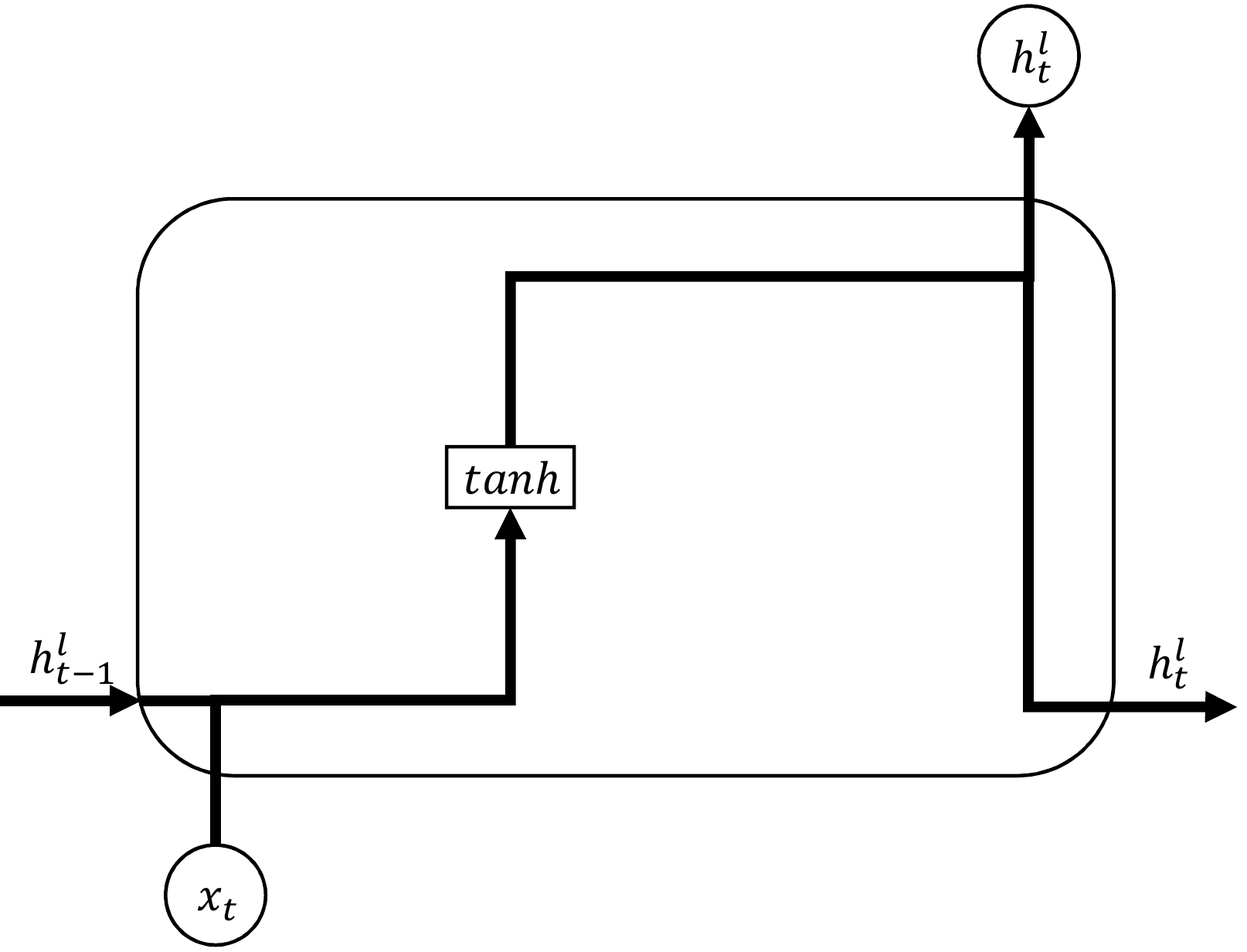}%
         \label{fig:traditional RNN cell}}\hspace{0.02\linewidth}
         \subfloat[LSTM cell]{\includegraphics[width=0.48\linewidth]{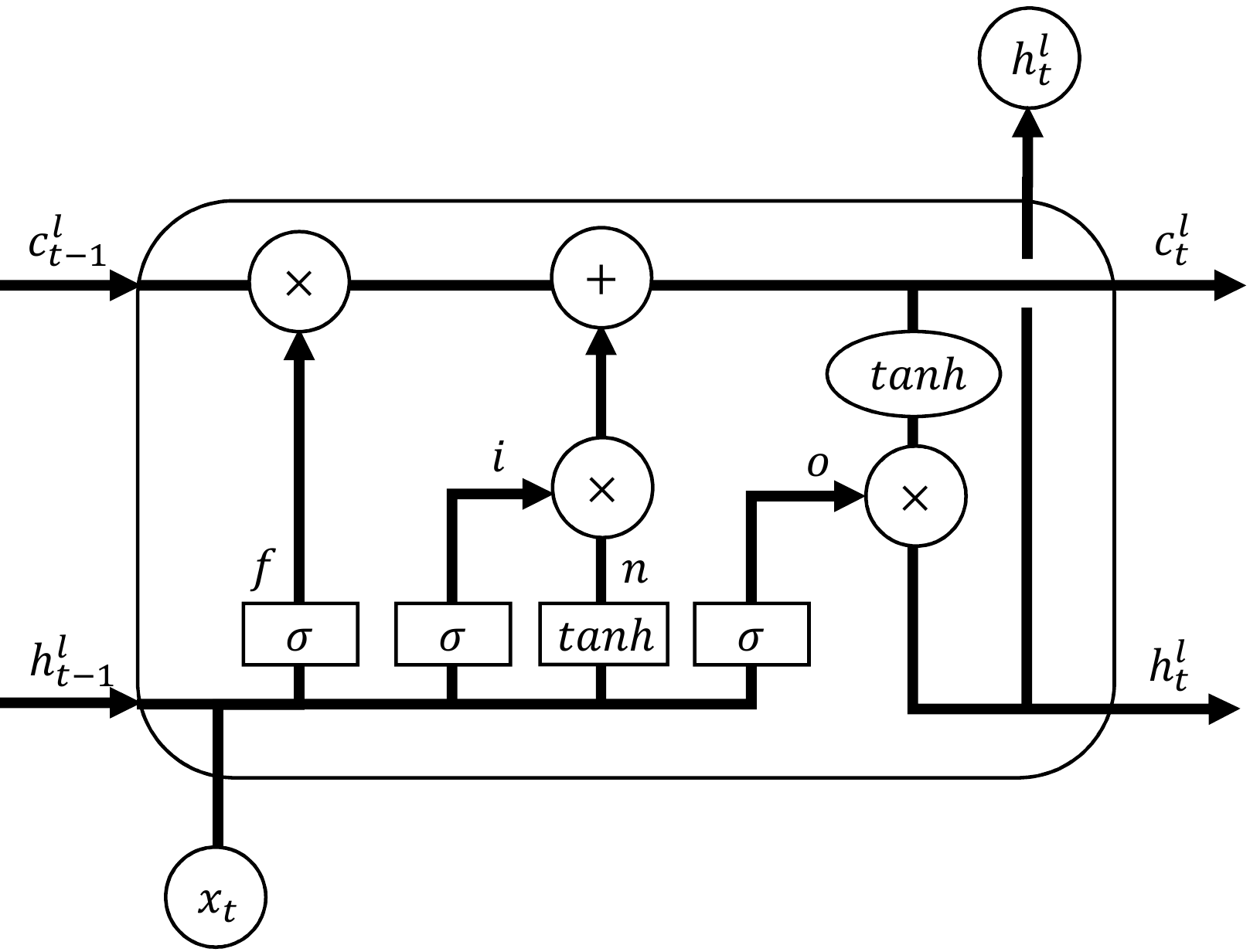}%
         \label{fig:lstm cell}}
    \caption{RNN cell}
    \label{fig:Cell}
\end{figure}

Nevertheless, the basic RNN is unable to learn the semantic dependency within longer time steps. LSTM~(Long short-term memory)~\cite{HochSchm97, zaremba2014recurrent} and GRU~(Gated recurrent unit)~\cite{cho2014learning} networks are generally deployed solutions, bringing gate mechanisms to RNN cell. Fig.~\ref{fig:Cell} provides the general RNN cell and LSTM cell as the example, in which $f, i, n, o$ stand for various gates\footnote{Gates of GRU are different but similar in design, not listed here.}, and $\sigma$ for sigmoid function. 
Unlike the plain RNN cell using a single tanh function to transfer the data, LSTM cell relies on cell states $c$~(special of LSTM networks) to maintain the context, to which multiple gates could remove or add information, and finally to produce the hidden state outputs. 

\subsection{Limitations of existing coverage metrics}\label{subsec: limitations}
While numerous coverage metrics~\cite{pei2017deepxplore, ma2018deepgauge, sun2018testing} are proposed for DNN testing, they are mostly based on CNN neurons and hard to satisfy RNN testing. First, an RNN usually comprises two or three layers each with several cells when unfolded, much fewer than a CNN usually of ten more layers each with hundreds of neurons. Second, sigmoid and tanh are conventionally used for an RNN cell whereas RELU is the most choice for a CNN neuron. This is critical because that the value range of RELU is $[0,\infty)$ while $[0,1]$ for sigmoid and $[-1,1]$ for tanh, leading to infinite values of CNN neurons whereas narrowed value ranges of RNN states.

Unfortunately, neuron-based coverage metrics fail to consider these critical characteristics. We evaluate neuron coverage~\cite{pei2017deepxplore} on our tested models, which treats the hidden states of each RNN cell as the equivalent output of a CNN neuron. As for the PTB language model comprising two layers each with 10 time steps, there will be only 20 neurons in total. The neuron coverage reaches 100\% with at most 4 inputs even taking a higher threshold 0.5. When for adversarial testing on this model, it fails to find any adversarial inputs, shown in Table \ref{table: RQ2 cov} of \S~\ref{sec:experiment}. 
As for coverage criteria ~\cite{ma2018deepgauge, ma2018combinatorial} defined over multi-section~(e.g. 1000) neuron value ranges discriminated by training and testing inputs, the narrow value ranges of RNN states~(hidden state value not exceeding $\pm 1$) limit their applications to RNN testing. 

Recently, DeepStellar~\cite{du2019deepstellar} abstracts RNN models as Discrete-Time Markov Chain (DTMC) models and then adapts coverage metrics of~\cite{ma2018deepgauge} for testing. Another work testRNN~\cite{huang2019coverage} designs novel coverage metrics for LSTM models to quantify temporal relations in RNNs. As to these coverage criteria, they primarily measure over abnormal values in testing, which are recognized by thresholds set according to training data. In this paper, we define coverage metrics trying to capture key inference logics inspired by distinctive roles of core RNN components, with no aid of training data.



%% file: sec-coverage.tex
\section{State Coverage Metrics}\label{sec:coverage}
In this section, we propose two state coverage metrics based on unique features of RNNs. Hidden state coverage~($HS\_C$) is designed to capture RNN prediction logics, which could be universally applied to RNN models. Due to the widespread deployments of LSTM networks, cell state coverage~($CS\_C$) is specially designed for LSTM models.


\subsection{Hidden State Coverage}
\begin{figure}[!htbp]
	\centering
	\subfloat[Hidden state]{\includegraphics[width=0.48\linewidth]{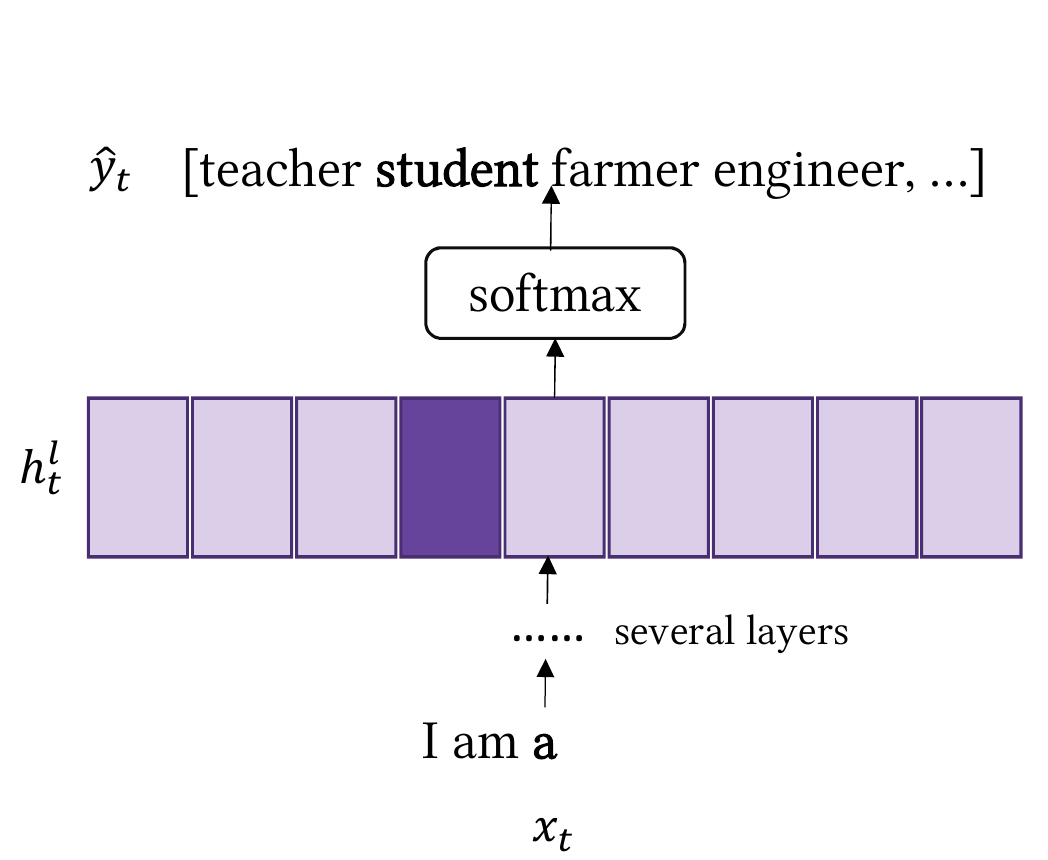}%
		\label{fig:hs_c}}%
	\subfloat[Cell state]{\includegraphics[width=0.48\linewidth]{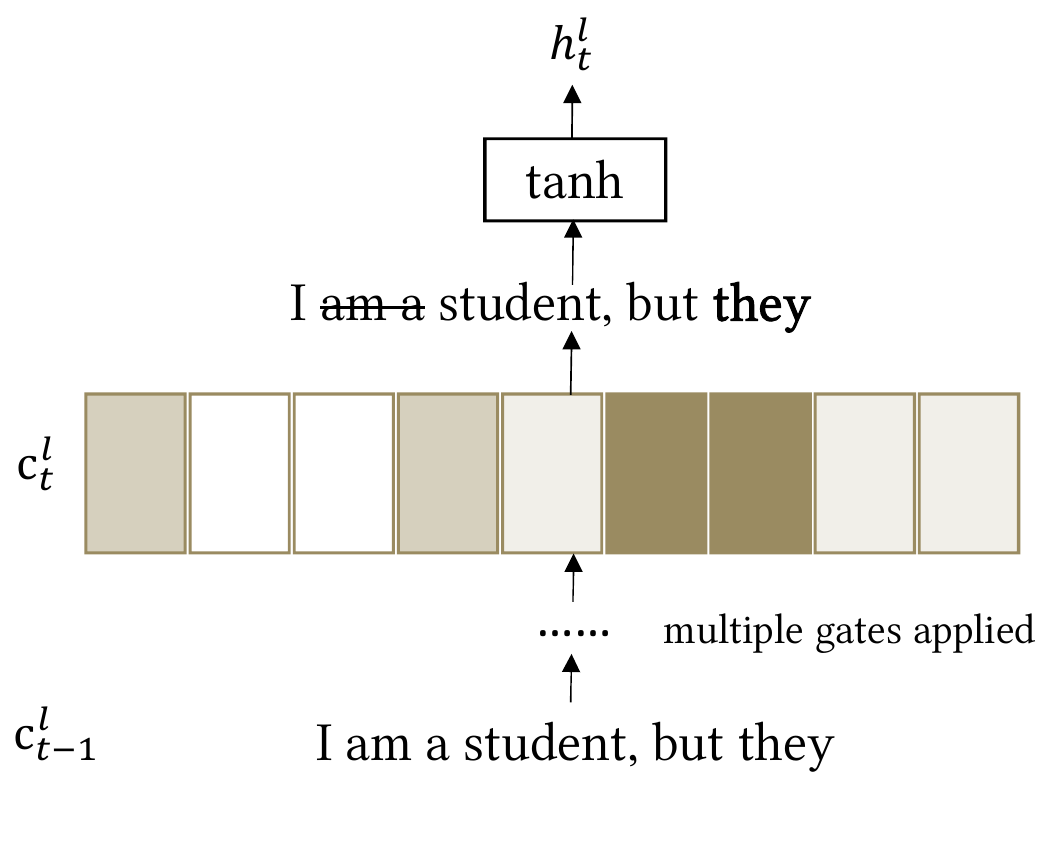}%
		\label{fig:cs_c}}
	\vspace{-0.1cm}
	\caption{Illustrations of RNN states.} 
	\label{fig:state Illustration}
\end{figure}

As discussed in \S~\ref{subsec: limitations}, RNN states cannot be regarded to be identical to CNN neurons. Fig.~\ref{fig:state Illustration} provides a simple illustration of the inner logics of hidden states and cell states. In Fig.~\ref{fig:hs_c}, $h_{t}^{l}$ represents the output of each RNN cell, which is a vector containing hundreds or thousands of hidden states. Here each rectangle represents each hidden state, where a darker color means a higher value. When to predict the next word following ``a", these hidden states will be mapped to a list of candidates. If a hidden state is the maximum, its mapped candidate will probably be the prediction result. 

\begin{figure*}[h]
	\centering
	\setlength{\belowcaptionskip}{-3pt}
	\includegraphics[width=0.9\linewidth]{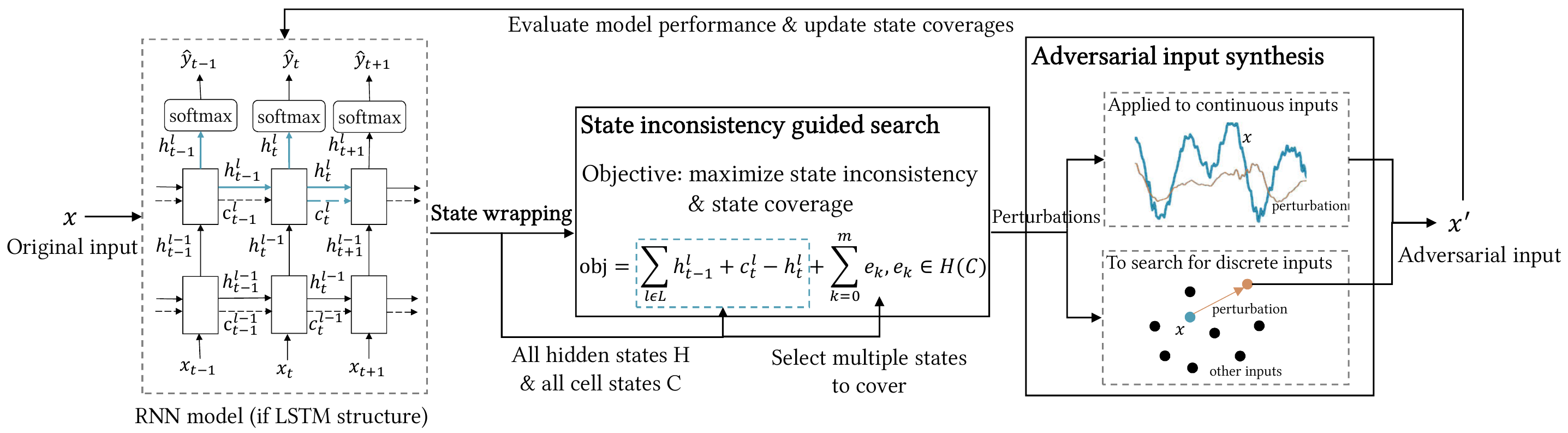}
	\vspace{-0.2cm}
	\caption{Architecture of RNN-Test}
	\label{fig:architecture}
\end{figure*}

Therefore, hidden states of each vector $h$ lead to varied prediction results for each step. To explore the inference logics thoroughly, it is meaningful for each hidden state to be the maximum in the vector and perform the key inference. We define hidden state coverage as the ratio of such hidden states of all the hidden states during testing. Formally, the definition is given in formula \eqref{eq:hidden state coverage}.

\begin{equation}\label{eq:hidden state coverage}
HS\_C = \frac{\left | \left \{ e \mid \forall e \in h, \forall h \in H, e=max(h) \right \} \right |}{\left | H \right |} 
\end{equation}

Assume $H$ denotes all the hidden states of an RNN model of given inputs, which is a four-dimensional matrix of shape $(T, L, B, E)$, where $T, L, B$ are the number of the time steps\footnote{For models could be fed with inputs of non-equal steps, $T$ will be adjusted according to the length of each input.}, layers and batch, respectively. $E$ is the state size. Though $H$ varies among RNN models, $T, L, B, E$ are always the necessary components, where batch is to accelerate computations by feeding multiple inputs simultaneously. Thus, $|H|$ will be $T\times L\times B\times E$. Here, a specific hidden state vector $h\in H$ contains $E$ hidden states and is denoted as its index of $H$, which is $[t, l, b]$, where $t\in\{1, 2, \ldots, T\}, l\in\{1, 2, \ldots, L\}, b\in\{1, 2, \ldots, B\}$. That means, $h$ is the output of the RNN cell of the $t$-th step in the $l$-th layer for the $b$-th input. For a state $e\in h$, $e$ is covered if its value $e = max(h)$. 

\subsection{Cell State Coverage}
As in Fig.~\ref{fig:lstm cell}, the cell states and gates are activated by functions sigmoid and tanh. The sigmoid function of three gates outputs values between 0 and 1, determining how much of each cell state to keep. The tanh function pushes the states between -1 and 1, for gate $n$ to add information to cell states, and for cell states $c$ to compute the hidden states. Thus, each cell state value protects the contexts. Fig.~\ref{fig:cs_c} illustrates the cell states, where $c_{t}^{l}$ is the output of each LSTM cell which is also a vector of cell states. Similarly, a rectangle is also a cell state and a darker color for a higher value. When to predict the predicate after ``they”, $c_{t}^{l}$ will receive contexts from $c_{t-1}^{l}$ to keep key semantics and remove those invalid, and then to compute predictions.

In this paper, we design cell state coverage over different value ranges standing for degrees to keep contexts. In the experiments, cell state values mostly fall into the central range while few be the boundary value. We suppose that covering more of each section~(5 sections in this paper) , especially boundary sections, could explore more context space. 
The formal definition is given in formula \eqref{eq:cell state coverage}. 

\begin{equation}\label{eq:cell state coverage}
CS\_C_{sec_{i}} = \frac{\left | \left \{ e \mid \forall e\in c, \forall c \in C, \tanh(e)\in sec_{i} \right \} \right |}{\left | C \right | }
\end{equation}

Here, all the cell states of an RNN model fed with given inputs are denoted as $C$, which is also a matrix of shape $(T, L, B, E)$. The value range of function tanh is split to $Sec$ sections and each section is $sec_{i} = [v_{i-1}, v_{i}]$, where $-1 \leq  v_{i} \leq 1$. For a specific cell state vector $c\in C$, it is also denoted by the index of $C$ as $[t, l, b]$. If a cell state $e\in c$ and its activation value $\tanh(e)\in sec_{i}$, $i\in \{1, 2, ..., Sec\}$, then $e$ is covered in $sec_{i}$. 

Thereby, we could measure how extensively the test inputs exercise RNN logics and benefit adversarial testing with state coverage metrics as guidance, without additional resources extracted from training data. 

%% file: sec-approach.tex
\section{RNN-Test Design}\label{sec:approach}

In this section, we present a technical description of RNN-Test in detail. Fig.~\ref{fig:architecture} depicts the overall workflow. Given the tested model, RNN-Test will focus on the kernel sequential part of the RNN structure without other components for particular tasks. For original test input $x$, we first extract the hidden states and cell states of each RNN cell by state wrapping, without affecting its inherent process. These states are crucial for the subsequent state inconsistency guided search, maximizing state inconsistency and state coverage to generate adversarial inputs. Unlike the usual idea of increasing the model cost~\cite{Goodfellow2015Explaining, gong2017crafting} or the probabilities of targeted classes~\cite{moosavi2016deepfool, guo2018dlfuzz}, RNN-Test boosts the inconsistency of RNN states opposite to their inner dependencies~(The part of $obj$ rounded with blue dashed frame violates data dependencies marked with blue lines in the model). In this way, RNN-Test could search for adversarial inputs in a lightweight and scalable means. Meanwhile, RNN-Test also tries to cover more states and explore more inference logics during testing, guided with specific state coverage information for different models. Then, the joint optimization problem will be solved in a gradient-based manner and acquire minute perturbations. 

Once obtained the perturbations, adversarial inputs are easy to acquire for models with continuous inputs like speech, by applying perturbations directly to original inputs. For models with discrete inputs like NLP tasks, the perturbation applied to the test input probably will not lead to a legal input. Here, we adopt the nearest one as the adversarial input after iteratively scaling the perturbation, thus avoiding the invalid input. 
Finally, these adversarial inputs will be assessed concerning the tested model for the performance and coverage, to improve subsequent testing efficiency. The detailed descriptions for each step are given below.
\subsection{State wrapping}
In the inherent implementation of an RNN model, there are two data structures accessible in the inference: all the hidden states of the last layer, and all the hidden states and cell states~(if LSTM underlying) of the last time step. For RNN testing, exploiting all the states should be a better choice for thoroughly searching for adversarial inputs. Therefore, we wrap the RNN cell implementation and keep all the hidden states and cell states of RNN cells in every layer and time step. With straightforward configurations, state wrapping will not interfere inner computation of the tested models.
Note that this step is not expensive and will not affect the time efficiency, with an open-source library~(20 lines of Python code). It is based on fundamental ``RNNCell”, the parent class of various cell
implementation, making it possible to be generalized to most RNN models. 

\subsection{State inconsistency guided search}
The state inconsistency guided search is the core portion of RNN-Test, an optimization problem composed of two parts.
It is formulated in equation \eqref{eq:state diff}, where the first part~($obj_{1}$) is referred to as adversary search, and the second part~($obj_{2}$) is as coverage guidance.
In addition, the two parts are free to be united together or alone, or even substituted with those of other methodologies, thus offering multiple possibilities of discovering adversarial inputs. 



\begin{equation}\label{eq:state diff}
\centering
\begin{cases}
 obj=obj_{1} + obj_{2} \\
 obj_{1} = \sum_{l\in L} {h_{t-1}^{l} + c_{t}^{l} - h_{t}^{l}} \\
 obj_{2} = \sum_{k=0}^{m} e_{k},\: e_k \in H(C)
\end{cases}
\end{equation}


\textbf{Adversary search}.
In RNN-Test, a novel methodology is designed to craft adversarial inputs specially for RNN models. 
As illustrated in Fig.~\ref{fig:Cell}, hidden state vector $h_{t}^{l}$ has a positive correlation with the inputs $h_{t-1}^{l}$, $h_{t}^{l-1}$ and intermediate outputs $c_{t}^{l}$, if $c$ implemented. Here, RNN-Test tries to increase $h_{t-1}^{l}$ and $c_{t}^{l}$ while decrease $h_{t}^{l}$ simultaneously, which intentionally violates their inner dependencies to lead the model to exhibit unusual behaviors. Then the violated dependencies will 
spread across the whole model. Thus, RNN-Test is able to search for adversarial inputs distributed outside of the regular inference space.

As for the time step $t$ in the objective, one step selected randomly out of each input will be adequate to achieve considerable performance. For the model with inputs always of hundreds of time steps, several more steps can be employed to increase the state inconsistency. Moreover, states of multiple layers $l\in L$~($L$ for all the layers) with respect to the same time step $t$ will be leveraged to accelerate the search efficiency.

\textbf{Coverage guidance}. This part aims to cover the uncovered states, exercising more decision logics to produce adversarial inputs. RNN-Test leverages the proposed $HS\_C$ and $CS\_C$ metrics to guide adversarial testing, where $CS\_C$ guidance for LSTM models and $HS\_C$ for general RNN models. To boost the specific coverage, RNN-Test selects $m$ hidden states or cell states to compose the optimization objective, as in formula~\eqref{eq:state diff}. Rather than merely  selecting uncovered states randomly, RNN-Test mainly chooses the states with values near to be covered so as to reach a higher coverage value at an earlier stage. Since $CS\_C$ is defined over a series of sections and the boundary sections are hardly covered, the states with values near the boundary section endpoints will be the targets to cover, thus leading RNN-Test to search in more sensitive space.  

Then, the joint optimization objective will be maximized by mutating the test inputs, unlike the training course minimizing the prediction error by tuning the parameters. Given the objective, its derivative for the input $x$ will be the perturbation, which is the gradient direction along which it increases or decreases most. Afterwards, the perturbations will be exploited to generate adversarial inputs.


\begin{table*}[h]
\centering
\setlength{\abovecaptionskip}{1pt} 
\caption{Summary of RNN models to evaluate RNN-Test. The first three models with sequential outputs are for major evaluation on sequential domains. The last model with classification outputs is constructed for comparison to testRNN.}
\setlength\tabcolsep{8pt}
\label{table: models}
\begin{threeparttable}
\begin{tabular}{c|c|c|c|c|c}
\hline
\multirow{2}{*}{\textbf{Model}} & \multirow{2}{*}{\textbf{Description}} & \multirow{2}{*}{\textbf{Architecture}} & \multicolumn{3}{c}{\textbf{Performance}} \\ \cline{4-6} 
 &  &  & Metric & Reported & Ours \\ \hline
\textbf{\begin{tabular}[c]{@{}c@{}}PTB language \\ model\end{tabular}} & \begin{tabular}[c]{@{}c@{}}General language \\ model\end{tabular} & \begin{tabular}[c]{@{}c@{}}Two-layer LSTM in its \\ small configuration(i.e. fewer steps)\end{tabular} & \begin{tabular}[c]{@{}c@{}}Train perplexity\tnote{1}\\ Test perplexity\end{tabular} & \begin{tabular}[c]{@{}c@{}}37.99\\ 115.91\end{tabular} & \begin{tabular}[c]{@{}c@{}}43.316\\ 117.122\end{tabular} \\ \hline
\textbf{\begin{tabular}[c]{@{}c@{}}Spell checker \\ model\end{tabular}} & \begin{tabular}[c]{@{}c@{}}Simple seq2seq\\ model\end{tabular} & \begin{tabular}[c]{@{}c@{}}Two-layer bi-direction\\ LSTM for the encoder\end{tabular} & Sequence loss & 15\% & 10\% \\ \hline
\textbf{\begin{tabular}[c]{@{}c@{}}DeepSpeech \\ ASR model\end{tabular}} & \begin{tabular}[c]{@{}c@{}}State-of-the-art\\ ASR model\end{tabular} & \begin{tabular}[c]{@{}c@{}}One-layer bi-direction\\ LSTM with CNN layers around\end{tabular} & WER\tnote{2} & 16\% & 16\% \\ \hline
\textbf{\begin{tabular}[c]{@{}c@{}}MNIST-LSTM \\ model\end{tabular}} & \begin{tabular}[c]{@{}c@{}}Handwritten digit\\recognition of LSTM network\end{tabular} & \begin{tabular}[c]{@{}c@{}}One-layer LSTM\\ \end{tabular} & Test accuracy & 98.3\% & 96.88\% \\ 
\hline
\end{tabular}
\begin{tablenotes}
\footnotesize
\item[1] Perplexity, the universal metric for language models, where lower perplexity corresponds to a better model.
\item[2] Word error rate, a common performance metric for seq2seq and ASR models, where higher WER means worse predictions.
\end{tablenotes}
\end{threeparttable}
\end{table*}

\begin{algorithm}[h]
	\caption{Adversarial input synthesis for discrete inputs}\label{alg:algorithm}
	\small
	\begin{algorithmic}[1]
		\Require $x$ $\leftarrow$ original test input
		\Statex \quad\ \ $t$ $\leftarrow$ one time step selected to modify
		\Statex \quad\ \ grad $\leftarrow$ perturbations obtained
		\Statex \quad\ \ embs $\leftarrow$ embeddings of the vocabulary
		\Statex \quad\ \ \footnotesize{MAX\_SCALE} \small $\leftarrow$ maximum degree of scaling the gradient
		\Statex \vspace{-5pt}\hspace{-15pt}\dotfill
		\State /*\textit{\footnotesize{generate adversarial inputs for NLP tasks.}}*/
		\Procedure{\footnotesize{GEN\_ADV}}{$x$, $t$, grad, embs}
		\State $x^{\prime}$ $\leftarrow$ $x$
		\State dist\_vec $\leftarrow$ $\emptyset$
		\For{scale $\in$ [1, \footnotesize{MAX\_SCALE}\small]}  \Comment{\textit{\footnotesize{search along the gradient}}}
		\State pert = grad$_{t}$ $\times$ scale \footnotesize{\Comment{\textit{perturbation for the time step}}}
		\State t\_emb = $x_{t}$ + pert \Comment{\textit{\footnotesize{get invalid embedding by gradient ascent}}} 
		\For{emb $\in$ embs} 
		\State dist = norm(t\_emb - emb) \Comment{\textit{\footnotesize{distance of t\_emb to emb}}}
		\State dist\_vec = dist\_vec $\cup$ $\{$dist$\}$ 
		\EndFor
		\vspace{-2pt}
		\State nearest\_emb = argmin(dist\_vec) \Comment{\textit{\footnotesize{the nearest embedding}}}
		\If{nearest\_emb != $x_{t}$} 
		\State $x_{t}^{\prime}$ = nearest\_emb \Comment{\textit{\footnotesize{modify the time step}}}
		\State \textbf{break} 
		\EndIf
		\EndFor
		\State \textbf{return} $x^{\prime}$ \Comment{\textit{\footnotesize{acquire the adversarial input}}}
		\EndProcedure
	\end{algorithmic}
\end{algorithm}

\subsection{Adversarial input synthesis}  
For continuous inputs like speech, the perturbations could be applied directly to acquire the adversarial input. For NLP tasks whose inputs are words or characters scattered in discrete embedding space, procedure GEN\_ADV is presented in Algorithm \ref{alg:algorithm}. In the procedure, we iteratively scale the gradient to be applied as the perturbation and then search for the nearest word/character in the embedding space to mutate the input step~(lines 8 to 14 in Algorithm \ref{alg:algorithm}).
This is a straightforward way to obtain valid adversarial inputs, ridding of embeddings which are not equivalent to legal words/characters. Besides, the embedding representations of words or characters in each NLP task are acquired after enough training, which could unveil their semantic properties. Therefore, searching along the gradient for the nearest embedding could get desired adversarial inputs with existing semantic information. 


In the literature, the adversarial input is identified for the imperceptibility from the original input but with the distinct class label. In sequential domains with no classifications, it is hard to recognize a generated sequence as the adversarial input avoiding false-positives, which has no standards yet~\cite{papernot2016crafting, odena2018tensorfuzz, carlini2018audio}. Fortunately, model performance metrics are a good choice to exhibit qualities of adversarial inputs, which are supposed to be accessible in all the tasks. Consequently, adversarial inputs obtained will be fed into the model assessing whether to decay the performance and updating the coverage, where coverage information will be exploited to guide subsequent testing.



%% file: sec-experiment.tex
\section{Experiment}\label{sec:experiment} 
\subsection{Experiment Setup}\label{sec:exp setup}
\textbf{Implementation.} We developed the framework RNN-Test on the widely deployed framework Tensorflow 1.3.0, and evaluated RNN-Test on a computer having Ubuntu 16.04 as the host OS, with an Intel i7-7700HQ@3.6GHz processor of 8 cores, 16GB of memory and an NVIDIA GTX 1070 GPU. 

As for hyperparameters in RNN-Test algorithms, such as $m$ and MAX\_SCALE, they are tuned for each tested model and not listed here for simplicity. We will release our code and datasets upon publication for further discussions.

\vspace{0.1cm}
\noindent\textbf{Tested models.} A summary of four tested models is presented in Table~\ref{table: models}. We mainly evaluated RNN-Test on the first three RNN models dealing with different sequential tasks. The last model of classification task is particularly constructed for comparison to testRNN. These models of common structures and various tasks provide more confidence for the generalization of RNN-Test to other RNN models. 

\textit{PTB language model}~\cite{ptb} is a well-known RNN model, basically to generate subsequent texts taking previous texts as input. It is the implementation of the fundamental LSTM~\cite{zaremba2014recurrent} without particular adaptations for specific applications. The training and testing data are provided by the Penn Tree Bank dataset~\cite{taylor2003penn}, and we extracted the first 25 sentences of the testing data for evaluation. We trained this model to achieve comparable performance to that reported using the same training course. 

\textit{Spell checker model}~\cite{spellchecker} is one of the widespread seq2seq models in NLP tasks, which receives a sentence with spelling mistakes as input and outputs the sentence with mistakes corrected. The training data are twenty popular books from project Gutenberg~\cite{Gutenberg}. For testing, we constructed 160 sentences with spelling mistakes like examples of developers, thanks to rich sources from Tatoeba~\cite{sentences}. Since their pre-trained model is unavailable, we trained this model in the same way.

\textit{DeepSpeech ASR model}~\cite{hannun2014deep} is a state-of-the-art speech-to-text RNN model employed in lots of security-critical scenarios. Its pre-trained model DeepSpeech-0.1.1~(Mozilla's implementation) could be deployed conveniently, and our testing data are the first 20 samples extracted from the Common Voice corpus~\cite{Commonvoice}. 

\vspace{0.1cm}
\noindent\textbf{Baselines.}
For comparison, we customize adversarial testing methodologies FGSM~\cite{Goodfellow2015Explaining} and DLFuzz~\cite{guo2018dlfuzz} to work for RNN models. They both generate adversarial inputs by solving optimization problems in a gradient-based manner. We implement their optimization objectives and coverage metrics on RNNs, while other procedures are the same as RNN-Test. For FGSM-based method, its optimization objective only contains the adversary search part without coverage guidance, whereas DLFuzz-based methodology also makes use of coverage guidance to obtain adversarial inputs, where neuron coverage~(NC) is the underlying metric. Here, the NC definition of RNN models is the same as DeepTest~\cite{tian2018deeptest}. Note that the customization will not degrade their performance since our optimized searching procedures are also used for them. 

For relevant works on RNN testing, there is a significant gap to conduct comparisons due to framework incompatibility. Tested models of testRNN~\cite{huang2019coverage} and DeepStellar~\cite{du2019deepstellar} are built on Keras and corresponding models on TensorFlow are mostly unavailable. Ultimately, we developed RNN-Test on an LSTM network~\cite{mnist-lstm} of MNIST dataset, which is an image classifier both evaluated in the two works. This MNIST-LSTM network is constructed on TensorFlow and achieves comparable test accuracy over the default MNIST dataset. 

\vspace{0.1cm}
\noindent\textbf{Research questions}: We constructed experiments to answer the following research questions.  
\begin{itemize}
\item \textbf{RQ1.} How is the effectiveness of the RNN-Test? (\S~\ref{subsec:RQ1})
\item \textbf{RQ2.} How is the effectiveness of coverage guidance for adversarial testing? (\S~\ref{subsec:RQ2})
\item \textbf{RQ3.} How is the quality of adversarial inputs obtained by RNN-Test? (\S~\ref{subsec:RQ3})
\end{itemize}

\begin{table*}[!htbp]
	\centering
	\setlength{\abovecaptionskip}{5pt}
	\caption{Effectiveness of RNN-Test and other methods in their default settings, measured over adversarial inputs obtained by each method. Note that worse performance values~(e.g. WER) indicate stronger test capability of methods~(The best result across each row is denoted bold). The coverage guidance used by RNN-Test is given following w.~(with), $HS\_C$ is hidden state coverage and $CS\_C$ is cell state coverage. The same symbols are used in below tables.} 
	\label{table: RQ1}
	\renewcommand{\arraystretch}{1.1}
	\begin{threeparttable}
		\begin{tabular}{c|c||c|ccc|>{\centering\arraybackslash}p{2cm}>{\centering\arraybackslash}p{2cm}}
			\hline
			\multicolumn{2}{c||}{} & \multicolumn{6}{c}{\textbf{Methodology}} \\ \hline
			\textbf{Model} & Performance  & \begin{tabular}[c]{@{}c@{}}Original\end{tabular}
			& \begin{tabular}[c]{@{}c@{}}Random testing\end{tabular} & \begin{tabular}[c]{@{}c@{}}FGSM-based\end{tabular} & \begin{tabular}[c]{@{}c@{}}DLFuzz-based\end{tabular} & \begin{tabular}[c]{@{}c@{}}RNN-Test\\ (w. $HS\_C$)\end{tabular} & \begin{tabular}[c]{@{}c@{}}RNN-Test\\ (w. $CS\_C$)\end{tabular} \\ \hline 
			\multirow{2}{*}{\textbf{\begin{tabular}[c]{@{}c@{}}PTB language\\ model\end{tabular}}} & Perplexity & 150.46 & 229.97 & 240.07 & 233.35 & \textbf{285.13} & 277.44 \\ 
			& Generation Rate\tnote{1} & - & \textbf{100.00\%} & 95.78\% & 93.59\% & \textbf{100.00\%} & \textbf{100.00\%} \\ \hline 
			\multirow{3}{*}{\textbf{\begin{tabular}[c]{@{}c@{}}Spell checker\\ model\end{tabular}}} & WER & 5.63 & 7.10 & 7.19 & 7.07 & 7.40 & \textbf{7.49} \\ 
			& BLEU\tnote{2} & 0.870 & 0.830 & 0.829 & 0.826 & 0.827 & \textbf{0.822} \\ 
			& Success Rate\tnote{3} & - & 64.58\% & 73.61\% & 73.61\% & 73.61\% & \textbf{76.39\%} \\ \hline 
			\multirow{3}{*}{\textbf{\begin{tabular}[c]{@{}c@{}}DeepSpeech\\ ASR model\end{tabular}}} & WER & 5.50 & 5.35 & 6.65   & 6.18   & \textbf{8.10}   & 7.80 \\ 
			& BLEU & 0.796  & 0.800 & 0.747  &  0.786  & \textbf{0.703}  & 0.720  \\ 
			& Success Rate & - & 40.67\% & 90.00\%   & 67.50\% & \textbf{100.00\%} & \textbf{100.00\%}
			\\ \hline
		\end{tabular}
		\begin{tablenotes}
			\footnotesize
			\item[1] Generation rate. Ratio of the test set the methodology has managed to produce the adversarial input.
			\item[2] BLEU~(Bilingual evaluation understudy). Correspondence of prediction and the ground truth, where higher BLEU means better predictions.
			\item[3] Success Rate. Ratio of the generated adversarial inputs to successfully reduce the model performance, not used for the first model as its performance is recorded over all inputs. 
		\end{tablenotes}
	\end{threeparttable}
\end{table*}

\begin{table}[h]
	\centering
	\setlength{\abovecaptionskip}{3pt}
	\caption{Effectiveness of RNN-Test compared to testRNN on MNIST-LSTM model over 500 original inputs. The results are listed in a similar way of testRNN~\cite{huang2019coverage}, where those for testRNN are exactly that they reported. } 
	\renewcommand{\arraystretch}{1.1}
	\label{table: RQ1 to testRNN}
	\begin{tabular}{c||cc|cc}
		\hline
		& \multicolumn{4}{c}{\textbf{Methodology}}                                                                                                                                                                                                              \\ \hline
		\begin{tabular}[c]{@{}c@{}}\textbf{MNIST-LSTM}\\ \textbf{model} \end{tabular}		       & \begin{tabular}[c]{@{}c@{}}testRNN\\ (RM)\end{tabular} & \begin{tabular}[c]{@{}c@{}}testRNN\\ (TM)\end{tabular} & \begin{tabular}[c]{@{}c@{}}RNN-Test\\ (w. $HS\_C$)\end{tabular} & \begin{tabular}[c]{@{}c@{}}RNN-Test\\ (w. $CS\_C$)\end{tabular} \\ \hline
		\begin{tabular}[c]{@{}c@{}}Test Cases\\ Generated\end{tabular}    & 2000                                                   & 2000                                                   & 500                                                          & 500                                                          \\ \hline
		\# Adv. Inputs          & 26                                                     & 63                                                     & \textbf{348}                                                          & 279                                                          \\ \hline
		\begin{tabular}[c]{@{}c@{}}Avg. Perturb.\\ (L2 norm)\end{tabular} & \textbf{1.051}                                                  & 4.028                                                  & 1.74                                                         & 1.69                                                         \\ \hline
		Adversary Rate            & 1.3\%                                                  & 3.15\%                                                 & \textbf{69.6\%}                                                       & 55.8\%                                                       \\ \hline
	\end{tabular}
\end{table}

\subsection{Effectiveness of RNN-Test (RQ1)}\label{subsec:RQ1}
To conduct a thorough evaluation, we compare our RNN-Test approach with other methodologies over the tested models, measuring the model performance fed with adversarial inputs obtained by each methodology. 
Besides, we also provide results of original test inputs, and those of random testing that randomly replaces a word/character of text input or applies Gaussian noise to speech input. We run them on each tested model over the same original test set three times, to alleviate the uncertainty each time. The same settings are adopted for below evaluations of other methods. 

\vspace{0.1cm}
\noindent\textbf{Overall results.} Table \ref{table: RQ1} summarizes the overall results, from which we could derive the following inferences. 
Firstly, adversarial inputs can decline the model performance, since tested models all achieve worse performance over adversarial input sets than the original test sets. 

Secondly, random testing methods can also obtain adversarial inputs, but they are far from satisfactory. For the 100\% generation rate on PTB language model, random replacement could always get mutated inputs while FGSM- and DLFuzz-based methods may fail to find adversarial inputs for some inputs. 

Thirdly, the RNN-Test approach outperforms FGSM-based and DLFuzz-based approaches, with more performance reduction and higher success~(or generation) rate. 
For instance, in comparison with FGSM-based method, RNN-Test~(w. $CS\_C$) achieves 15.57\% higher perplexity and 4.22\% higher generation rate on PTB language model, 4.17\% higher WER, 0.84\% lower BLEU and 2.78\% higher success rate on spell checker model, and 17.29\% higher WER, 3.61\% lower BLEU and 10\% higher success rate on DeepSpeech ASR model. As for the slight improvement on the spell checker model, the sparse embedding space may be the primary cause, since it largely limits the searching capability.

\vspace{0.1cm}
\noindent\textbf{How to choose the appropriate coverage guidance.}  
As shown in Table \ref{table: RQ1}, RNN-Test guided with $HS\_C$ and $CS\_C$ both gain better effectiveness than other methodologies, with no one always superior to the other. When applied for realistic RNN tasks, it is straightforward to choose the appropriate coverage guidance. For LSTM models, both are good alternatives. For common RNN and GRU models, $HS\_C$ is the choice, since hidden states are universal across these structures. 

\vspace{0.1cm}
\noindent\textbf{Comparison to relevant RNN testing works.} 
As described in \S~\ref{sec:exp setup}, we conduct comparisons to other RNN testing methods over one MNIST-LSTM model due to framework incompatibility. 
Table~\ref{table: RQ1 to testRNN} presents the comparison of RNN-Test to testRNN on MNIST-LSTM model, both over 500 original inputs as the evaluation setting of testRNN. TestRNN employed random mutation~(RM) and targeted mutation~(TM) to generate 2000 test cases for evaluation, respectively. In RNN-Test, once one adversarial input is obtained for the corresponding test input, the testing procedure starts for another test input, and thus acquired 500 test cases. 
In Table~\ref{table: RQ1 to testRNN}, RNN-Test could generate much more adversarial inputs out of fewer test cases, obtaining 52.65\% to 66.45\% higher adversary rate than TM algorithm of testRNN, a refined means of RM. Although testRNN~(RM)
obtains smallest perturbations, testRNN~(TM) with largest perturbations still reached limited adversary rate.
From another point of view, RNN-Test is also competitive to be applied to classification domains. 

As to DeepStellar, it reported that thousands of adversarial inputs were obtained for 100 test inputs after 6 hours on a high-performance server. Despite our design, RNN-Test is also able to test each input continuously and generate much more adversarial inputs. If with a same high-performance server~(with 28-core CPU, 196 GB RAM, and 4 NVIDIA Tesla V100 16G GPUs), RNN-Test is supposed to achieve comparable results.

\begin{figure*}[h]
	\centering
	\setlength{\belowcaptionskip}{-5pt}
	\subfloat[PTB language model]{\includegraphics[width=0.295\linewidth]{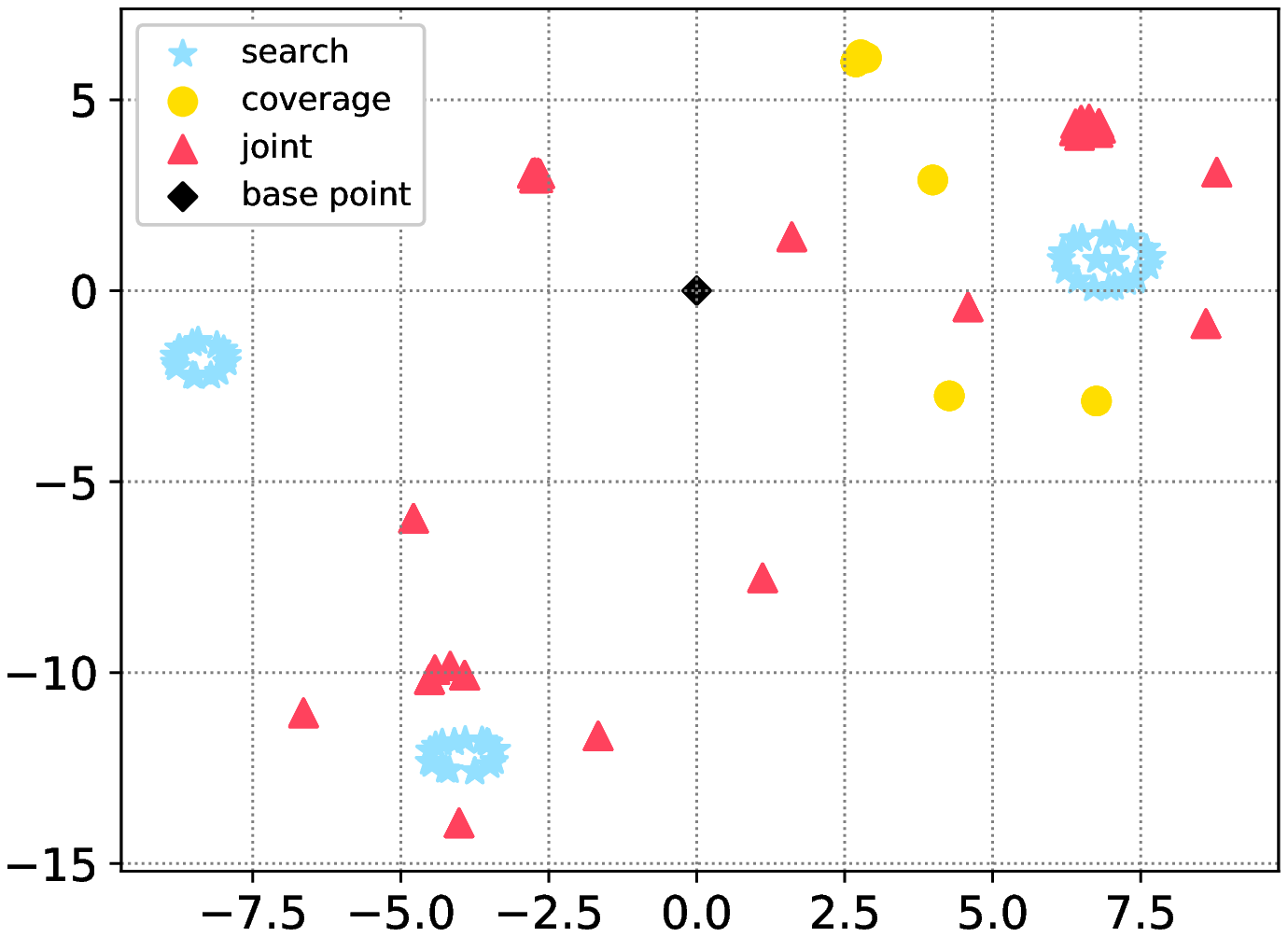}%
		\label{fig:ptb pert}}
	\subfloat[spell checker model]{\includegraphics[width=0.29\linewidth]{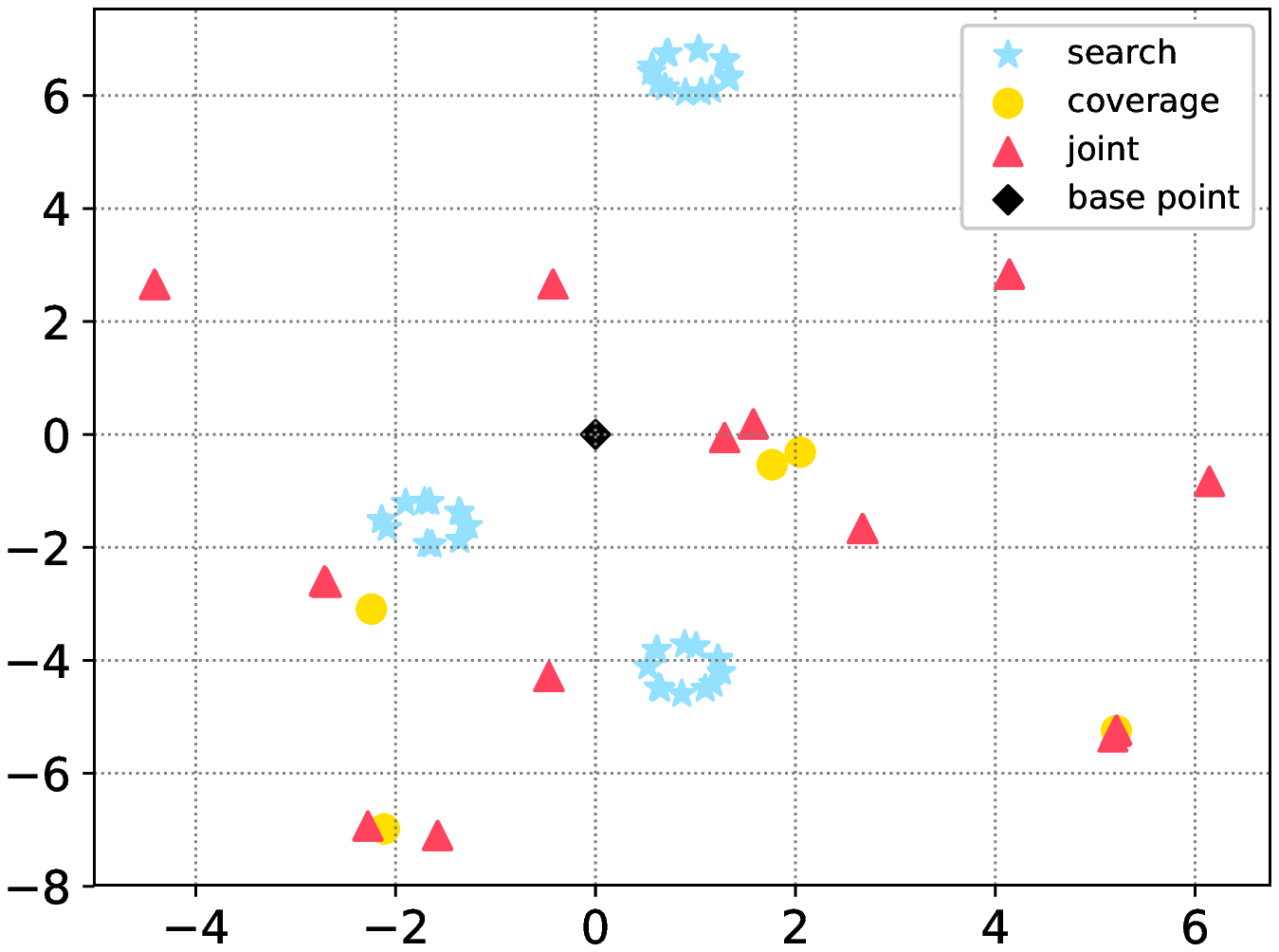}%
		\label{fig:sp pert}}
	\subfloat[DeepSpeech model]{\includegraphics[width=0.29\linewidth]{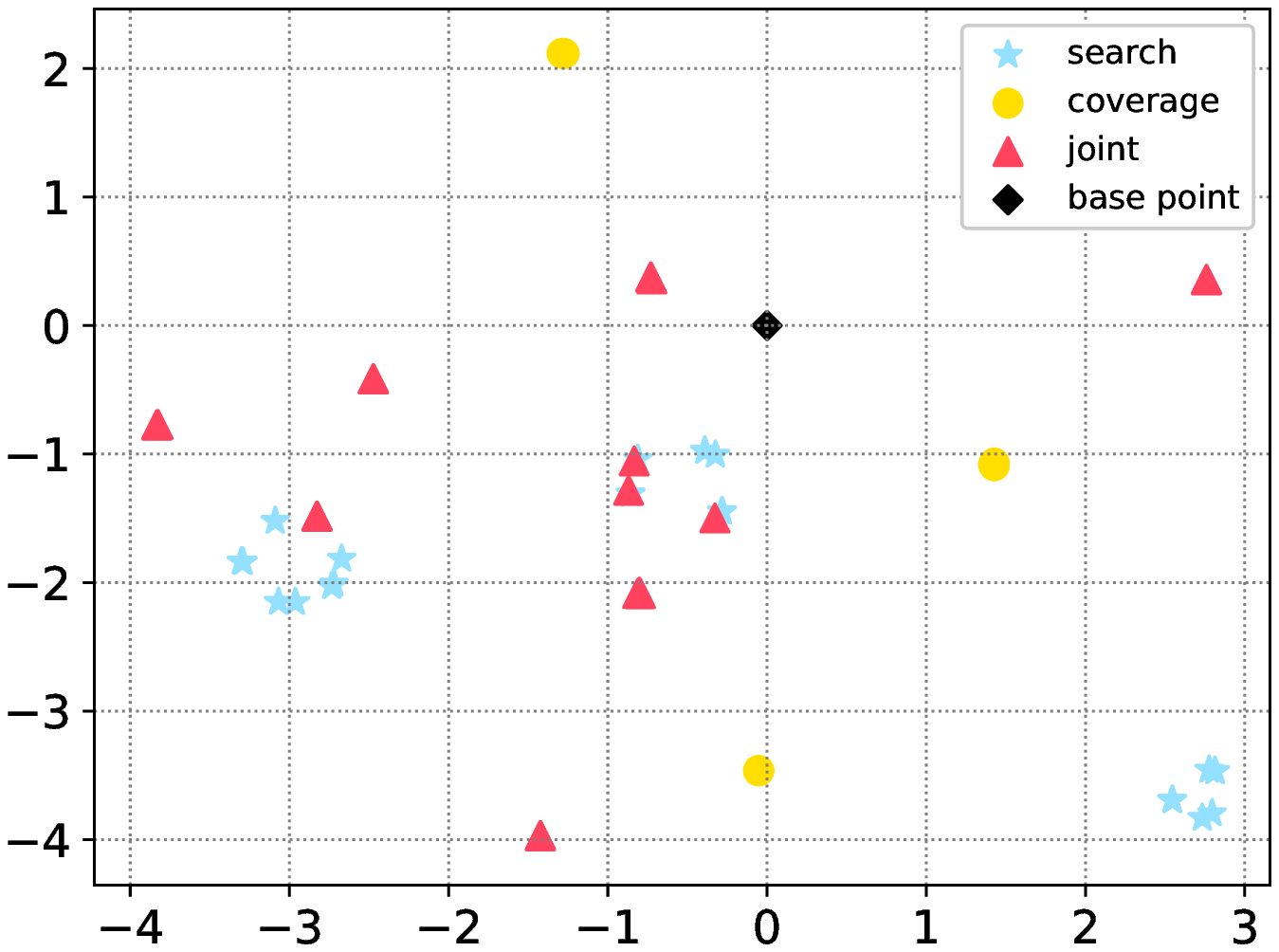}%
		\label{fig:audio pert}}
	\vspace{-0.1cm}
	\caption{TSNE transformations of perturbations of the above approaches with different optimization objectives for one same test input. Search is adversary search, coverage is coverage guidance, joint is joint objective, same in below figures. The divergent distribution represents various perturbations and thus adversarial inputs.}
	\label{fig:pert}
\end{figure*}

\begin{tcolorbox}[width=\linewidth, halign=justify, colframe=black, colback=gray!10, boxsep=1mm, boxrule = 0.3mm, arc=2mm, left = 0.5mm, right = 0.1mm, top = 0.3mm, bottom = 0.3mm]
\quad The answer to RQ1: The RNN-Test approach is effective in generating adversarial inputs, with the ability to reduce the model performance sharply with high success~(or generation) rate.
\end{tcolorbox}

\subsection{State coverage guidance contributes to adversarial testing~(RQ2)}\label{subsec:RQ2} 
\textbf{Divergent perturbations of coverage guidance.}
The previous work~\cite{li2019misleading} suggests that perturbations obtained by neuron coverage guidance are similar to adversary-based search methods~(e.g. FGSM) and so the coverage guidance does not add too much, which is concluded based upon analyses over popular coverage guided testing methodologies~\cite{pei2017deepxplore, ma2018deepgauge, ma2018combinatorial}. But the conclusion may not work for the proposed state coverage metrics, since those criteria assessed are all over CNN neurons. 

Here, we recorded perturbation vectors obtained by approaches we evaluated over the same inputs of each RNN model. Besides their default settings, we run each approach with either pure adversary search or coverage guidance, as well as the joint way. To visualize, we leverage the state-of-the-art high-dimensional reduction technique TSNE~\cite{maaten2008visualizing} to transform multi-dimensional perturbation vectors into two dimensions. 
As Fig.~\ref{fig:pert} shows, there is no evident similarity of perturbations of the adversary search, coverage guidance or joint objectives. In contrast, the divergent distribution implies that coverage guidance is capable to offer alternative perturbations, whether utilized independently or jointly, thus providing varied adversarial inputs. Therefore, the coverage guidance is worthy to be applied to adversarial testing. 


\textbf{Effectiveness of pure coverage guidance for adversarial testing}. 
State coverage guidance can also be adopted to discover adversarial inputs independently, due to the unique perturbations of coverage guidance.
Table~\ref{table: RQ2 cov} presents the results of $HS\_C$, $CS\_C$, and $NC$ as guidance to be the only optimization respectively. 
As shown, state coverage metrics as guidance can acquire adversarial inputs on the tested models, while $NC$ guidance fails to obtain any on PTB language model. Overall, both $CS\_C$ and $HS\_C$ outperform $NC$ as guidance, especially on PTB language model and DeepSpeech ASR model. Therefore, neuron-based coverage metrics will not be appropriate for RNN models, as discussed in \S~\ref{subsec: limitations}. 
Surprisingly, when cross-referenced with Table~\ref{table: RQ1} for the spell checker model,  $CS\_C$ guidance exhibits the best performance with the highest WER and success rate. 

\begin{table}[t]
	\centering
	\caption{Effectiveness of state coverage guidance, compared to neuron coverage guidance.}
	\vspace{-0.2cm}
	\label{table: RQ2 cov}
	\begin{tabular}{c|c||c|c|c}
		\hline
		\multicolumn{2}{c||}{}  & \multicolumn{3}{c}{\textbf{Methodology}} \\ \hline 
		\textbf{Model} & Performance & $NC$ & $HS\_C$ & $CS\_C$ \\ \hline 
		\multirow{2}{*}{\textbf{\begin{tabular}[c]{@{}c@{}}PTB language\\ model\end{tabular}}} & Perplexity & 150.46 & \textbf{238.07} & 236.96 \\ 
		& Generation Rate & 0\% & 94.66\% & \textbf{97.22}\% \\ \hline 
		\multirow{3}{*}{\textbf{\begin{tabular}[c]{@{}c@{}}Spell checker\\ model\end{tabular}}} & WER & 7.03 & 7.48 & \textbf{8.00} \\ 
		& BLEU & 0.830 & 0.827 & \textbf{0.823} \\ 
		& Success Rate & 73.61\% & 75.00\% & \textbf{77.78\%} \\ \hline 
		\multirow{3}{*}{\textbf{\begin{tabular}[c]{@{}c@{}}DeepSpeech\\ ASR model\end{tabular}}} & WER & 5.45 & \textbf{5.65} & 5.35 \\ 
		& BLEU & 0.801 & \textbf{0.791} & 0.794 \\ 
		& Success Rate & 10.00\% & 40.00\% & \textbf{55.00\%} \\ \hline
	\end{tabular}
\end{table}

\textbf{Enhancement of coverage guidance to other methods.} 
We also demonstrate that both FGSM-based and DLFuzz-based approaches with state coverage guidance could gain higher effectiveness than themselves and those jointed with $NC$ guidance. 
For instance, Table~\ref{table: RQ2 enh} provides results of DLFuzz-based methodology jointed with $HS\_C$ and $CS\_C$ guidance, as well as its $NC$ guidance. Compared with $NC$ guidance, $HS\_C$ and $CS\_C$ guidance improve DLFuzz-based technique over the tested models in varying degrees. Additionally, similar results are attained for FGSM-based approach, where state coverage guidance improves more than $NC$ guidance, as presented in Table~\ref{table: RQ2 enh cost}.
Thus, state coverage guidance is proven to be able to enhance other adversarial testing methodologies. Though these two methods could be improved with state coverage guidance, the most powerful means is still the RNN-Test, as cross-referenced with Table \ref{table: RQ1}. 

\begin{table}[h]
	\centering
	\caption{Effectiveness of DLFuzz-based methodology with state coverage guidance, compared to its $NC$ guidance.}
	\vspace{-0.2cm}
	\label{table: RQ2 enh}
	\setlength\tabcolsep{5pt}
	\begin{tabular}{c|c||c|c|c}
		\hline
		\multicolumn{2}{c||}{} & \multicolumn{3}{c}{\begin{tabular}[c]{@{}c@{}}\textbf{Methodology}\\(DLFuzz-based)\end{tabular}} \\ \hline
		\textbf{Model} & Performance & w. $NC$ & w. $HS\_C$ & w. $CS\_C$\\ \hline
		\multirow{2}{*}{\textbf{\begin{tabular}[c]{@{}c@{}}PTB language\\ model\end{tabular}}} & Perplexity & 233.35 & \textbf{243.19} & 238.71 \\ 
		& Generation Rate & 95.42\% & 97.44\% & \textbf{99.15\%} \\ \hline
		\multirow{3}{*}{\textbf{\begin{tabular}[c]{@{}c@{}}Spell checker\\ model\end{tabular}}} & WER & 7.07 & \textbf{7.44} & 7.10 \\ 
		& BLEU & 0.826 & \textbf{0.825} & \textbf{0.825} \\ 
		& Success Rate & 73.61\% & 75.00\% & \textbf{76.39\%} \\ \hline
		\multirow{3}{*}{\textbf{\begin{tabular}[c]{@{}c@{}}DeepSpeech\\ ASR model\end{tabular}}} & WER & \textbf{6.18} & 6.15 & 6.10 \\ 
		& BLEU & 0.786 & 0.785 & \textbf{0.778} \\ 
		& Success Rate & 67.50\% & \textbf{75.00\%} & 70.00\% \\ \hline
	\end{tabular}
\end{table}

\begin{figure*}[h]
	\centering
	\subfloat[$CS\_C_{sec_{5}}$ on PTB language model]{\includegraphics[width=0.33\linewidth]{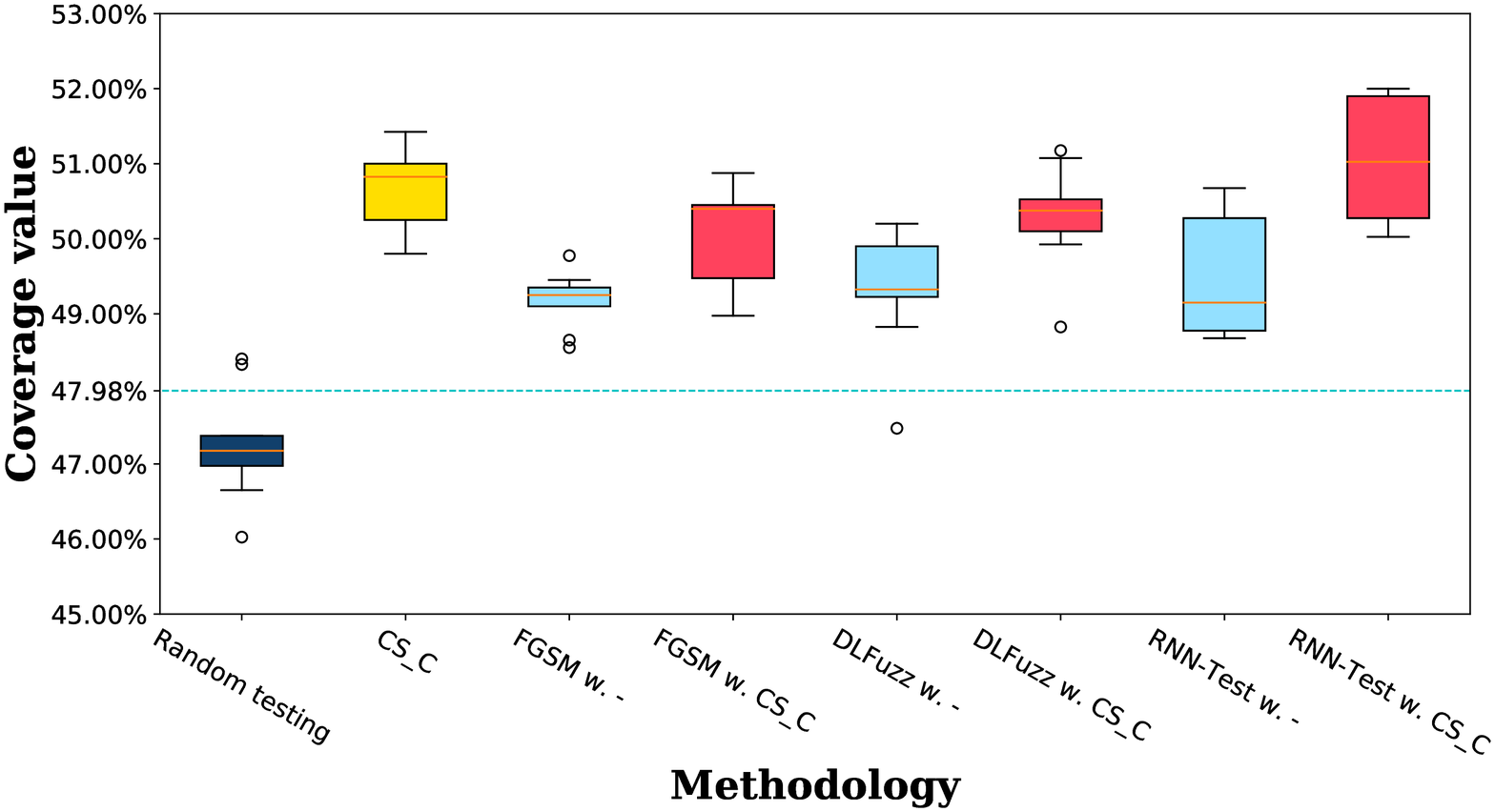}%
		\label{fig:ptb box}}
	\subfloat[$NC$ on spell checker model]{\includegraphics[width=0.33\linewidth]{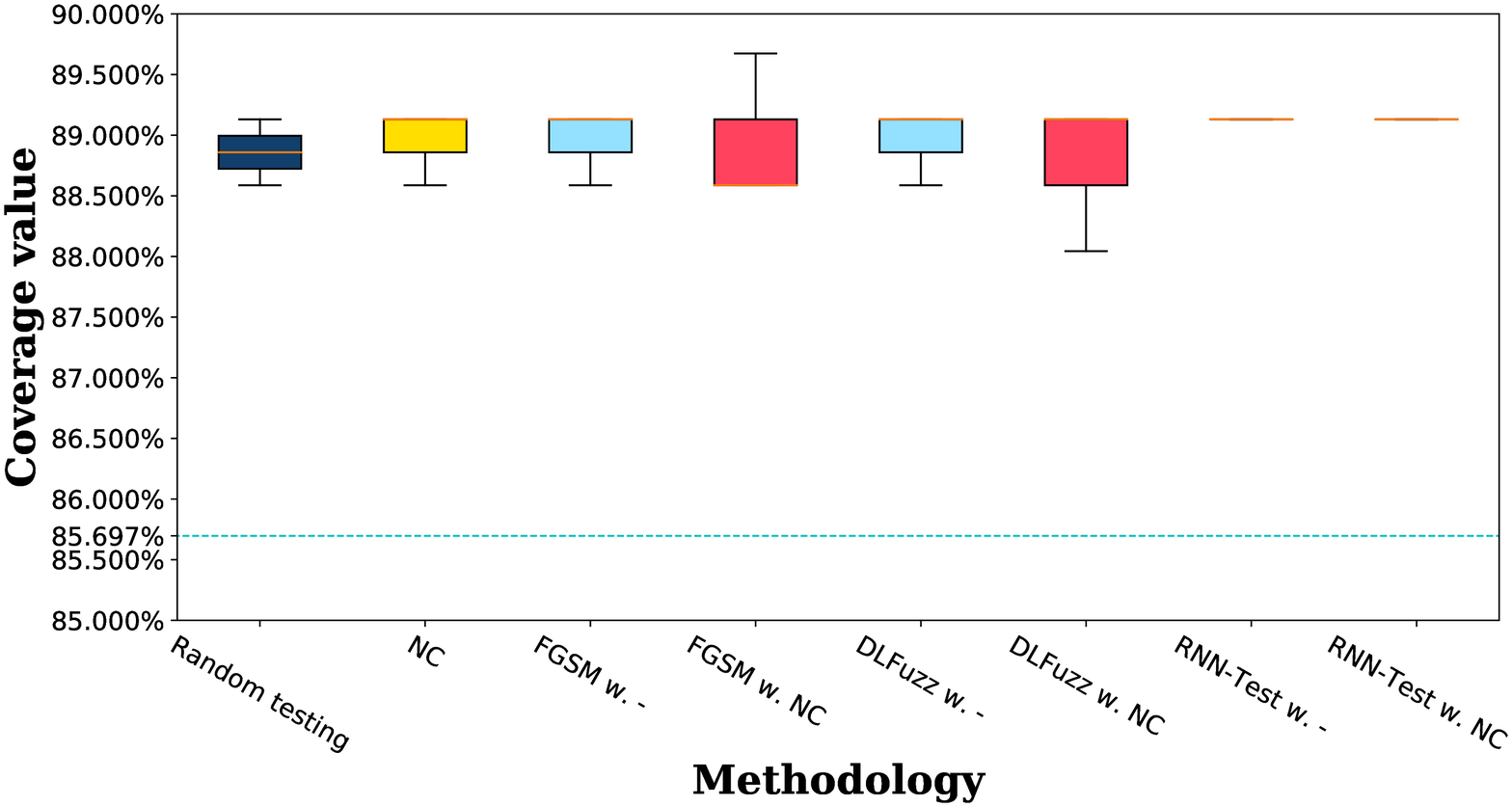}%
		\label{fig:sp box}}
	\subfloat[$HS\_C$ on DeepSpeech ASR model]
	{\includegraphics[width=0.33\linewidth]{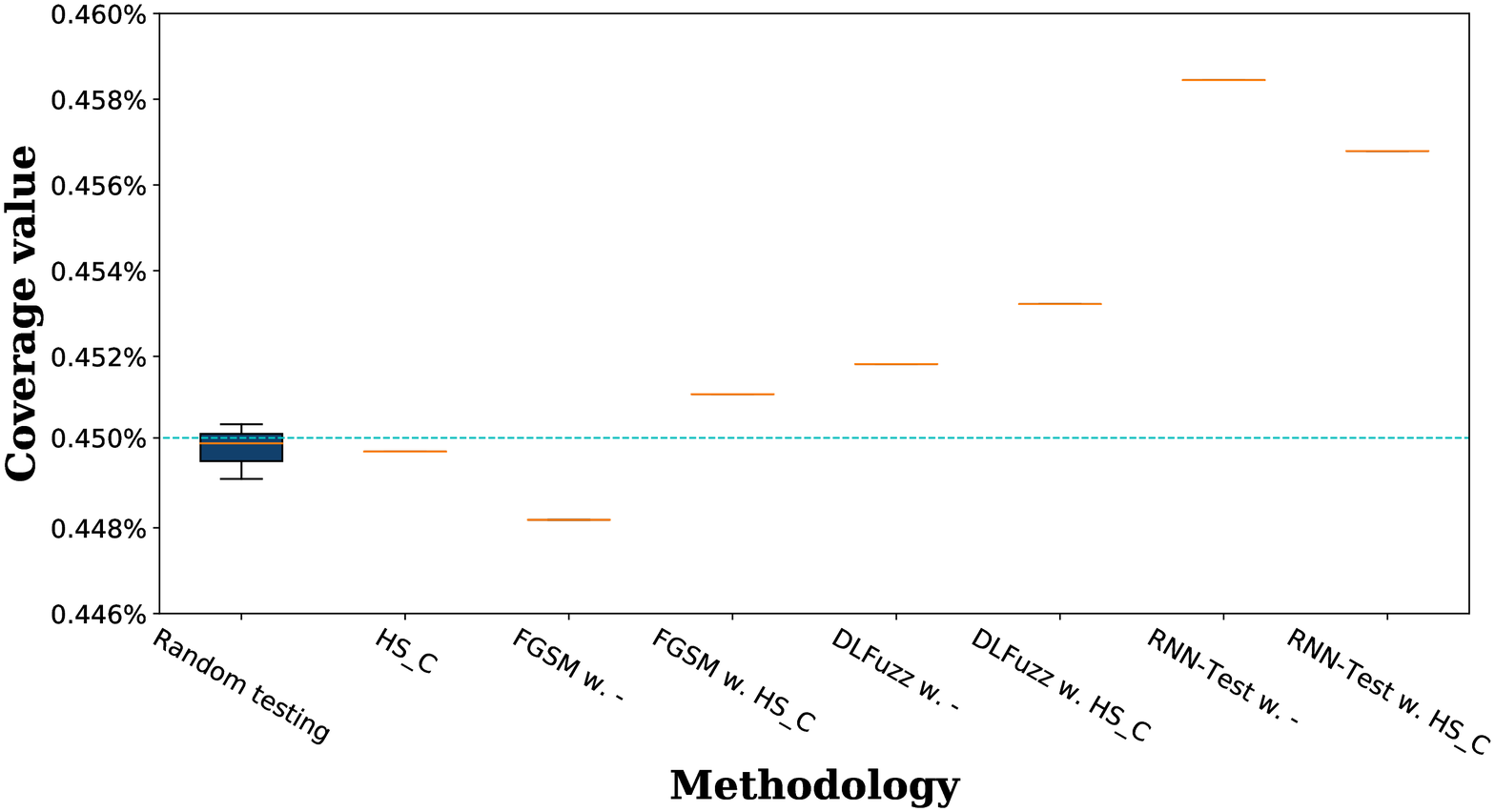}%
		\label{fig:audio box}}    
	\vspace{-2pt}
	\caption{Value ranges of coverage metrics among different approaches~(w.- denotes no coverage guidance) over the same amount of adversarial inputs. The blue dashed lines denote the corresponding coverage value of original test input sets.} 
	\label{fig:box}
\end{figure*}


\begin{table}[h]
	\centering
	\caption{Effectiveness of FGSM-based methodology jointed with the coverage metrics, compared to its default setting with no coverage guidance.}
	\setlength{\abovecaptionskip}{5pt} 
	\label{table: RQ2 enh cost}
	\setlength\tabcolsep{2pt}
	\begin{tabular}{c|c||c|c|c|c}
		\hline
		\multicolumn{2}{c||}{} &  \multicolumn{4}{c}{\begin{tabular}[c]{@{}c@{}}\textbf{Methodology}\\(FGSM-based)\end{tabular}}  \\ \hline
		\textbf{Model} & Performance & w. - &  w. $NC$ & w. $HS\_C$ & w. $CS\_C$ \\ \hline
		\multirow{2}{*}{\textbf{\begin{tabular}[c]{@{}c@{}}PTB language\\ model\end{tabular}}} & Perplexity & 240.07 & 241.23 & \textbf{256.91} & \textbf{256.91} \\ 
		& Generation Rate & 95.78\% & 98.69\% & 97.65\% & \textbf{100.00\%} \\ \hline
		\multirow{3}{*}{\textbf{\begin{tabular}[c]{@{}c@{}}Spell checker\\ model\end{tabular}}} & WER & 7.19 & 6.99 & \textbf{7.46} & 7.14 \\ 
		& BLEU & 0.829 & 0.832 & \textbf{0.828} & 0.831 \\ 
		& Success Rate & \textbf{73.61\%} & \textbf{73.61\%} & \textbf{73.61\%} & 70.83\% \\ \hline
		\multirow{3}{*}{\textbf{\begin{tabular}[c]{@{}c@{}}DeepSpeech\\ ASR model\end{tabular}}} & WER & 6.65 & 6.70 & 6.75 & \textbf{6.80} \\ 
		& BLEU & 0.747 & 0.747 & 0.748 & \textbf{0.746} \\ 
		& Success Rate & \textbf{90.00\%} & \textbf{90.00\%} & \textbf{90.00\%} & \textbf{90.00\%} \\ \hline
	\end{tabular}
\end{table}
\vspace{-0.5cm}

\textbf{Coverage value may not be a strong indicator of methodology effectiveness.}
Numerous works~\cite{pei2017deepxplore, ma2018deepgauge, ma2018combinatorial, huang2019coverage} adopt the coverage value as an indicator of effectiveness for adversarial testing methodologies.
Meanwhile, researchers~\cite{li2019misleading, dong2019there} raised doubts that there may be limited correlations between coverage and robustness of DNNs.

Here, we have analyzed correlations between the model performance and values of coverage metrics on the first three models, but found out weak positive or even negative correlations~(not given for brevity). Therefore, we could not draw the conclusion that obtaining higher coverage definitely results in higher effectiveness. For example, higher coverage value does not indicate higher success rate or WER on DeepSpeech ASR model. Though testRNN proved that their adversarial input set is with higher coverage rate than the normal input set, whether a higher coverage rate leads to a higher adversary rate also remains unsettled.  

From the current point of view, we suggest that more efforts should be put into coverage guidance for adversarial testing, but not just to boost the coverage value. Indicators related to the model performance should be considered with more weight to assess an adversarial testing method.

\textbf{Simple illustration of value ranges of coverage metrics.}
Fig.~\ref{fig:box} presents the value ranges of $HS\_C$, $CS\_C$ and $NC$ achieved by different methods on each tested model respectively. We also provide results of each methodology with and without the corresponding coverage guidance, since the coverage guidance still tends to improve the value. For each box, it represents a set of coverage values of the methodology at different times, marked with bounds and the median. Note that boxes in Fig.~\ref{fig:audio box} resemble lines because of the limited stochasticity and equal coverage values obtained~(except random testing) on that model. 

It must be claimed that coverage values strongly depend on the number of test inputs, and the same amount of inputs are supposed to be with similar value ranges. 
As presented, the value ranges of these coverage metrics among methodologies vary not much~(within 5\%), especially $NC$ ranges are almost the same. It is the same case for figures not given here. 
Furthermore, $HS\_C$ values are always very low since it is inherently hard to boost $HS\_C$ with a few inputs, similar to boundary sections of $CS\_C$ over larger models. 
Meanwhile, 
it also supplies evidence that methodology effectiveness may be affected little by coverage values. However, coverage guidance is still worthy of more research investment.
In summary, we could get the following answer. 
%

\vspace{0.1cm}
\begin{tcolorbox}[width=\linewidth, halign=justify, colframe=black, colback=gray!10, boxsep=1mm, boxrule = 0.3mm, arc=2mm, left = 0.5mm, right = 0.1mm, top = 0.3mm, bottom = 0.3mm]
\quad The answer to RQ2: State coverage metrics as guidance are able to acquire adversarial inputs, superior to neuron coverage guidance whether independently or jointed with adversary search. The coverage guidance has the potential to be more effective, since the divergent perturbations and best performance on the spell checker model.
\end{tcolorbox}

\subsection{Quality of adversarial inputs of RNN-Test~(RQ3)}\label{subsec:RQ3}
\begin{figure}[h]
	\centering
	\includegraphics[width=0.45\textwidth]{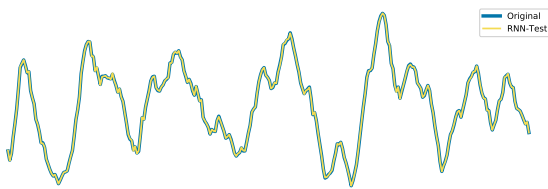}
	\vspace{-0.2cm}
	\caption{An example adversarial input of RNN-Test for DeepSpeech ASR model. The waveform of a test input~(blue, thick line) is overlapped with the waveform of the adversarial input~(yellow, thin line). Each waveform is 500 samples long and was chosen randomly from the corresponding inputs. The original prediction is ``the shop as closed on mondays" while the prediction for adversarial input is ``the shop as close tan monas". }
	\label{fig:audio wav}
\end{figure}

\textbf{Samples of adversarial inputs. }
Table~\ref{table: samples} lists samples of adversarial inputs on the two NLP models, with each approach to modify the same word. For both models, RNN-Test tends to generate different words with other methods, offering diverse adversarial inputs. For PTB language model, our adversarial inputs could result in the model sampling words farther from semantics and generating higher perplexity texts, where results of RNN-Test are with totally wrong semantics. Meanwhile, adversarial inputs for the spell checker model could result in the corrected mistakes in original inputs appearing again in the predictions of adversarial inputs. For DeepSpeech ASR model, they could result in the model making wrong predictions, as depicted in Fig.~\ref{fig:audio wav}. Such adversarial inputs of misleading semantics may harm the security requirements of tested models. 

\begin{table*}[h]
	\centering
	\caption{Samples of adversarial inputs on the tested models, the targeted words to modify are in red and underlined. The affected results for the spell checker model are also underlined.}
	\label{table: samples}
	\begin{tabular}{c|l|l}
		\hline
		\textbf{Methodology}                                         & \multicolumn{1}{c|}{\textbf{PTB language model}}                                                                                     & \multicolumn{1}{c}{\textbf{spell checker model}}                                                                                                                                                                                                                                                              \\ \hline
		Original                                                     & \begin{tabular}[c]{@{}l@{}}Input: no it was n't black {\color{red} \underline{Monday}} ...\\ Perplexity: 259.67 \\Generate: economy goes forward\\ \phantom{Generate: } on behalf of ...\end{tabular} & \begin{tabular}[c]{@{}l@{}}Input: I would swim through {\color{red} \underline{theoocean}} just to see your smile again.\\ Predict: I would swim through \underline{the ocean} just to see your smile again.\\ Input: The sound of yur voice {\color{red} \underline{islike}} siren's songto me.\\ Predict: The sound of yur voice \underline{is like} siren' song to me.\end{tabular}     \\ \hline \hline
		\begin{tabular}[c]{@{}c@{}}FGSM-based\end{tabular}    & \begin{tabular}[c]{@{}l@{}}Input: no it was n't black {\color{red} \underline{co.}} ...\\ Perplexity: 376.46\\Generate: economy goes forward\\ \phantom{Generate: }on behalf of ...\end{tabular}    & \begin{tabular}[c]{@{}l@{}}Input: I would swim through {\color{red} \underline{theootean}} just to see your smile again.\\ Predict: I would swim through \underline{the otean} just to see your smile again.\\ Input: The sound of yur voice {\color{red} \underline{isliee}} siren's songto me. \\ Predict: The sound of yur voice \underline{is liee} siren' song to me.\end{tabular}    \\ \hline
		\begin{tabular}[c]{@{}c@{}}DLFuzz-based\end{tabular} & \begin{tabular}[c]{@{}l@{}}Input: no it was n't black {\color{red} \underline{due}} ...\\ Perplexity: 357.38\\Generate: soviets appear reluctant\\ \phantom{Generate: } between france 's ...\end{tabular}    & \begin{tabular}[c]{@{}l@{}}Input: I would swim through {\color{red} \underline{theootean}} just to see your smile again. \\ Predict: I would swim through \underline{the otean} just to see your smile again.\\ Input: The sound of yur voice {\color{red} \underline{islske}} siren's songto me. \\ Predict: The sound of yur voice \underline{issle} siren' song to me.\end{tabular}     \\ \hline \hline
		\begin{tabular}[c]{@{}c@{}}RNN-Test\end{tabular}  & \begin{tabular}[c]{@{}l@{}}Input: no it was n't black {\color{red} \underline{\$}} ...\\ Perplexity: 513.91\\Generate: soviets appear reluctant\\ \phantom{Generate: } toward nov. a.m. ...\end{tabular}      & \begin{tabular}[c]{@{}l@{}}Input: I would swim through {\color{red} \underline{theoKcean}} just to see your smile again. \\ Predict: I would swim through \underline{the cocean} just to see your smile again.\\ Input: The sound of yur voice {\color{red} \underline{isltke}} siren's songto me. \\ Predict: The sound of yur voice \underline{istle} siren' song to me.\end{tabular}    \\ \hline
	\end{tabular}
\end{table*}

\textbf{Time efficiency, tiny perturbations and reality concerns. }
RNN-Test also has high time efficiency, producing each adversarial input costs 3s, 11s, and 24s on average on PTB language model, spell checker model, and DeepSpeech ASR model, respectively. That means RNN-Test has the potential to be applied in industrial practice.
In terms of the perturbations, only one word or character is modified for the former two models of text inputs. For DeepSpeech ASR model, the averaged perturbations are smaller than 0.04 in L2 norm. Therefore, adversarial inputs of RNN-Test appear benign to humans and ensure imperceptibility. 
As for realistic concerns about adversarial images of tasks like autonomous driving, adversarial cases will probably not encounter in real-world circumstances and weaken the security significance. In RNN testing, users are likely to mistype text inputs and fail NLP models in reality. Nevertheless, this is an urgent issue for ASR models at present, since adversarial audios~\cite{carlini2018audio} always become invalid when played over-the-air, which we will be devoted to in future works.



\textbf{Improve the model by retraining. }
Last but not the least, adversarial inputs obtained by RNN-Test are also capable to improve the model performance by retraining. We tried on PTB language model and incorporated adversarial inputs~(82.5 KB) to the training set~(5.1 MB). 

\begin{table}[!htbp]
	\centering
	\caption{The perplexity before and after retraining on PTB language model. Columns 3 and 5 are for the augmented training set. Columns 4 and 7 are for the improvement of retraining results w.r.t original results.}
	\label{table: RQ3}
	\setlength\tabcolsep{3pt}
	\setlength{\belowcaptionskip}{5pt}
	\begin{tabular}{c||c|c|c||c|c|c}
		\hline
		\multirow{2}{*}{\textbf{epoch}}& \multicolumn{3}{c||}{\textbf{train perplexity}} & \multicolumn{3}{c}{\textbf{valid perplexity}} \\ \cline{2-7}
		& original & w. adv. & increment & original & w. adv. & decrement \\ \hline
		0 & 290.584 & 288.579 & -0.690\% & 190.004 & 192.096 & -1.101\% \\ \hline
		2 & 113.216 & 113.712 & 0.439\% & 140.328 & 140.339 & -0.008\% \\ \hline
		4 & 86.290 & 87.195 & 1.049\% & 132.589 & 132.969 & -0.287\% \\ \hline
		6 & 56.282 & 56.961 & 1.207\% & 121.410 & 120.566 & 0.695\% \\ \hline
		8 & 46.549 & 47.082 & 1.146\% & 122.981 & 121.611 & 1.114\% \\ \hline
		10 & 43.991 & 44.474 & 1.096\% & 123.065 & 121.385 & 1.365\% \\ \hline
		12 & 43.227 & 43.695 & 1.082\% & 122.440 & 121.020 & 1.159\% \\ \hline
	\end{tabular}
\end{table}

Table~\ref{table: RQ3} presents the perplexity of PTB language model before and after retraining, where train perplexity indicates the performance on the training set while valid perplexity for the valid set. Here the data are averaged over 5 times of the same retraining process with 12 epochs, to mitigate affects due to the intrinsic indeterminism of neural networks. From columns 4 and 7, results show that the train perplexity of the model after retraining increases by 1.082\% whereas the valid perplexity decreases by 1.159\% in end. Moreover, the test perplexity after retraining is 102.75, which is also declined by 12.582\% compared to the original test perplexity 117.53.
Notice that even by incorporating fewer adversarial inputs~(1.6KB), the valid perplexity still declines by 0.058\%. 
Therefore, adversarial inputs could alleviate the over-fitting issue in training by reducing little train performance, but improving the valid and test performance and thus the robustness of RNN models. 

\vspace{0.1cm}
\begin{tcolorbox}[width=\linewidth, halign=justify, colframe=black, colback=gray!10, boxsep=1mm, boxrule = 0.3mm, arc=2mm, left = 0.5mm, right = 0.1mm, top = 0.3mm, bottom = 0.3mm]
	\quad The answer to RQ3: RNN-Test could efficiently produce adversarial inputs of high quality, declining the model performance sharply and improving the model by retraining. 
\end{tcolorbox}


%% file: sec-discussion.tex
\section{Threats to validity}\label{sec:threats}
Though RNN-Test exhibits appreciable effectiveness with the default setting in evaluations, its performance is inevitably influenced by the parameters, including the number of states selected to boost, the weights applied to joint objectives and the scaling degree of perturbations, especially the ways of sections splitting of $CS\_C$. They are worthy to be well explored in the future work. Furthermore, the uncertainty running each time still exists, owing to stochastic word/character to modify, which could be diminished by fixing the target. 

In addition, RNN-Test is devoted to being general and scalable for variants of RNNs, but we could not exhaustively evaluate all the variants and applications. In this paper, the structures of tested models are general to some extent, but training the spell checker model still costs hard work, due to its bad reproducibility of the training results given. Moreover, the state wrapping is designed to avoid interfering with model logics, but the adaptation efforts may be necessary for some variants with complex structures. 



%% file: sec-relatedwork.tex
\section{Related Work}\label{sec:relatedwork} 
\textbf{Adversarial deep learning.} 
The concept of adversarial attacks was first introduced in ~\cite{szegedy2013intriguing}. It discovered that state-of-the-art DNNs would misclassify the input images by applying imperceptible perturbations, where these mutated inputs are called adversarial examples/inputs. Their work FGSM~\cite{Goodfellow2015Explaining} and numerous following works~\cite{ moosavi2016deepfool, carlini2017towards,papernot2016limitations} generate adversarial inputs by maximizing the prediction error in a gradient-based manner. They provide rich input resources for CNNs to improve their robustness. 

Afterwards, \cite{papernot2016crafting} explains adversarial inputs for RNNs, but presents rough qualitative descriptions for those of sequential outputs. As mentioned before, existing works mostly focus on tasks of categorical outputs like sentiment analysis~\cite{samanta2017towards}. For those of text inputs, some works~\cite{samanta2017towards, sato2018interpretable} add, delete or substitute a word/character to construct adversarial inputs, leading models to give wrong classifications.  

Unfortunately, few works are evaluating the majority RNN models processing sequential outputs. Due to no explicit class labels and no standards for such adversarial inputs, present works attack these models to perform abnormally in various ways. TensorFuzz~\cite{odena2018tensorfuzz} crafts adversarial inputs to lead the language model to sample words from the blacklist. Several works~\cite{carlini2018audio} fool well-known ASR models to produce targeted phrases given by the attacker. In this paper, we propose RNN-Test as an effective and scalable methodology for diverse models of sequential outputs. 

\vspace{0.1cm}
\noindent\textbf{Coverage guided testing.} 
Based upon the exposed threats of DNNs, traditional software testing techniques are subsequently applied to test DNN systems, where coverage guided testing is of a popular trend. DeepXplore~\cite{pei2017deepxplore} first introduces neuron coverage which is defined over CNN neurons with pre-defined thresholds. Then, DeepGauge~\cite{ma2018deepgauge} defines a set of coverage metrics with finer-grained granularity, where neuron value ranges are split as thousands of sections according to training data. DeepCT~\cite{ma2018combinatorial} is even fine-grained to measure over combinations of neuron outputs. As stated in \S~\ref{subsec: limitations}, these neuron-based coverage metrics can not be applied to RNN states directly. 

As for works also among the first attempts of adversarial testing for RNN systems, DeepStellar~\cite{du2019deepstellar} adapts coverage metrics of DeepGauge to test RNN models, which need to be abstracted as a Markov Chain first. Despite its effectiveness, it is inevitable to miss key features and introduce computation overhead owing to intrinsic properties of abstraction. 
Another work testRNN~\cite{huang2019coverage} designs novel coverage metrics according to structures of LSTM models, some of which are special to quantify temporal relations. 
Besides of similar classification tasks, these two works both mutate inputs directly~(e.g. random noise) and use saturated coverage values to terminate testing. Our RNN-Test approach mutates inputs based on RNN logics and adopts coverage guidance to help search for adversarial inputs, which is effective for both sequential and classification domains.



%% file: sec-conclusion.tex
\section{Conclusions}\label{sec:conclusions}
We design and implement the adversarial testing framework RNN-Test for recurrent neural networks. 
RNN-Test focuses on testing the main sequential structure without limit to tasks, aggregating advantages of both the proposed search method and novel state coverage metrics as guidance. It is superior to existing methodologies for DNN testing and could effectively produce adversarial inputs over RNN models of various applications, reducing model performance evidently with high success~(or generation) rate. We also first demonstrate that coverage guidance has a diverse searching capability for adversarial inputs compared with other methods and our state coverage guidance outperforms neuron coverage guidance in RNN testing.
